\documentclass[11pt]{article}
\usepackage[margin=1in]{geometry}
\usepackage{cite}
\usepackage{amsmath,amssymb,amsfonts}
\usepackage{subcaption}
\usepackage{dsfont}
\usepackage{graphicx}
\usepackage{algorithm,algorithmic}
\usepackage{hyperref}
\hypersetup{hidelinks}
\usepackage{cuted}
\usepackage{textcomp}
\def\BibTeX{{\rm B\kern-.05em{\sc i\kern-.025em b}\kern-.08em
    T\kern-.1667em\lower.7ex\hbox{E}\kern-.125emX}}
\begin{document}
\title{Does DQN Learn?}

\author{Aditya Gopalan\thanks{Aditya Gopalan is with the Department of Electrical Communication Engineering, Indian Institute of Science, Bengaluru 560 012, India (email: aditya@iisc.ac.in).}
% \IEEEmembership{Senior Member, IEEE},
 \ and Gugan Thoppe\thanks{Gugan Thoppe is with the  Computer Science and Automation Dept., Indian Institute of Science (IISc), Bengaluru 560 012, India. He is also an Associate Researcher with the Robert Bosch Centre for Data Science and Artificial Intelligence (RBCDSAI), IIT Madras, Chennai 600 036, India (e-mail: gthoppe@iisc.ac.in).}}

\newcommand{\bmu}{\bar{\mu}}
\newcommand{\co}{\textnormal{co}}
\newcommand{\cl}{\textnormal{cl}}
\newcommand{\df}{\textnormal{d}}
\newcommand{\Lim}{\textnormal{Lim}}
\newcommand{\indc}{\mathds{1}}
\newcommand{\de}{d^{\epsilon}}
\newcommand{\De}{D^{\epsilon}}
\newcommand{\Pep}{P^{\epsilon'}}
\newcommand{\pie}{\pi^\epsilon}
\newcommand{\piep}{\pi^{\epsilon'}}
\newcommand{\brep}{\br^{\epsilon'}}
\newcommand{\tr}{T}
\newcommand{\vol}{\textnormal{vol}}
\newcommand{\stdbasis}{e}
\newcommand{\sgn}[1]{\text{sgn}\left( #1 \right)}
\newcommand{\intr}{\textnormal{int}}

\newcommand{\thetam}{\theta^{-}}
\newcommand{\thS}{\theta_*}
\newcommand{\piS}{\pi_*}
\newcommand{\QS}{Q^*}

\newcommand{\ba}{\mathbf{\bar{a}}}
\newcommand{\br}{\mathbf{r}}

\newcommand{\bE}{\mathbb{E}}
\newcommand{\bI}{\mathbb{I}}
\newcommand{\bP}{\mathbb{P}}
\newcommand{\bR}{\mathbb{R}}

\newcommand{\cA}{\mathcal{A}}
\newcommand{\cB}{\mathcal{B}}
\newcommand{\cC}{\mathcal{C}}
\newcommand{\cD}{\mathcal{D}}
\newcommand{\cF}{\mathcal{F}}
\newcommand{\cL}{\mathcal{L}}
\newcommand{\cP}{\mathcal{P}}
\newcommand{\cR}{\mathcal{R}}
\newcommand{\cS}{\mathcal{S}}
\newcommand{\cX}{\mathcal{X}}

\newcommand{\diag}{\textnormal{diag}}
\newcommand{\sign}{\textnormal{sgn}}
\newcommand{\hp}{h^\prime}

\newcommand{\red}[1]{{\color{red} #1}}
\newcommand{\blue}[1]{{\color{blue} #1}}
\newcommand{\done}{\hfill \blue{Done}.}

\renewcommand{\Pr}{\bP}

\newcommand{\ag}[1]{{\color{red} AG: #1}}
\newcommand{\gt}[1]{{\color{blue} GT: #1}}

\newcommand{\wip}[1]{{\color{red} #1}}

\newcommand{\landmark}{landmark}
\newcommand{\supp}{\textnormal{supp}}

\newcommand{\dataset}{{\cal D}}
\newcommand{\fracpartial}[2]{\frac{\partial #1}{\partial  #2}}

\newtheorem{theorem}{Theorem}
\newtheorem{remark}[theorem]{Remark}
\newtheorem{lemma}[theorem]{Lemma}
\newtheorem{proposition}[theorem]{Proposition}
\newtheorem{corollary}[theorem]{Corollary}

\allowdisplaybreaks        % in the preamble

\maketitle

\begin{abstract}
A primary requirement for any reinforcement learning method is that it should produce policies that improve upon the initial guess. In this work, we show that the widely used Deep Q-Network (DQN) fails to satisfy this minimal criterion---even when it gets to see all possible states and actions infinitely often (a condition under which tabular Q-learning is guaranteed to converge to the optimal Q-value function). Our specific contributions are twofold. First, we numerically show that DQN often returns a policy that performs worse than the initial one. Second, we offer a theoretical explanation for this phenomenon in \emph{linear DQN}, a simplified version of DQN that uses linear function approximation in place of neural networks while retaining the other key components  such as $\epsilon$-greedy exploration, experience replay, and target network. Using tools from differential inclusion theory, we prove that the limit points of linear DQN correspond to fixed points of projected Bellman operators. Crucially, we show that these fixed points need not relate to optimal---or even near-optimal---policies, thus explaining linear DQN's sub-optimal behaviors. We also give a scenario where linear DQN always identifies the worst policy. Our work fills a longstanding gap in understanding the convergence behaviors of Q-learning with function approximation and $\epsilon$-greedy exploration.
\end{abstract}

% \begin{IEEEkeywords}
% Deep reinforcement learning, Function approximation, Nonlinear dynamical systems, Reinforcement learning, Sliding mode control
% \end{IEEEkeywords}

\section{Introduction}
\label{s:Introduction}
Deep Q-Network (DQN) \cite{mnih2015human} is popular in Reinforcement Learning (RL) due to its groundbreaking success in mastering complex tasks, such as playing a video game. Notably, DQN has achieved human-level performance on a variety of Atari 2600 games, demonstrating its potential for learning and decision-making in environments with high-dimensional sensory inputs. This success stems from four innovations: (i) using a \textit{neural network} to reasonably approximate $\QS$---the optimal Q-value function---in the large state-action scenario, (ii) employing \textit{$\epsilon$-greedy policy} to balance exploration and exploitation of optimal actions at different states, (iii) leveraging \textit{experience replay} to decouple the algorithm's submodule that interacts with the environment from the one that updates $\QS$'s estimate, and (iv) incorporating a \textit{target network} to stabilize training. In recent times, though, the DQN algorithm has also been reported to exhibit several pathological behaviors (beyond the classical instability \cite{baird1995residual}) such as policy oscillation, i.e., alternating between two or more policies without end, and convergence to sub-optimal policies (including the worst) \cite{gordon1996chattering, gordon2000reinforcement, de2000existence, bertsekas2011approximate, young2020understanding}. In fact, Patterson et al. \cite{patterson2024empirical} claim the following:

\renewenvironment{quote}{%
  \list{}{%
    \leftmargin0.5cm   % this is the adjusting screw
    \rightmargin\leftmargin
  }
  \item\relax
}
{\endlist}

\begin{quote}
``  ... we observed (rare) catastrophic failure events for DQN across nearly every tested domain ... In Lunar Lander,
some agents would simply fly off into oblivion, obtaining incredible amounts of negative
reward until the episode was mercifully terminated ... In Cliff World, DQN would get stuck in a corner perpetually ... some agents would learn to jump into the cliff immediately to obtain massive negative rewards."
\end{quote}

These conflicting narratives lead us to three questions about DQN's behavior: (1) Does DQN ensure a monotonic improvement in the optimal policy estimate? (2) If not, does it at least ensure convergence to a locally optimal policy? (3) At the very least, is there any improvement over the initial policy? 

\begin{figure}
    \centering
    \includegraphics[width=0.5\linewidth]{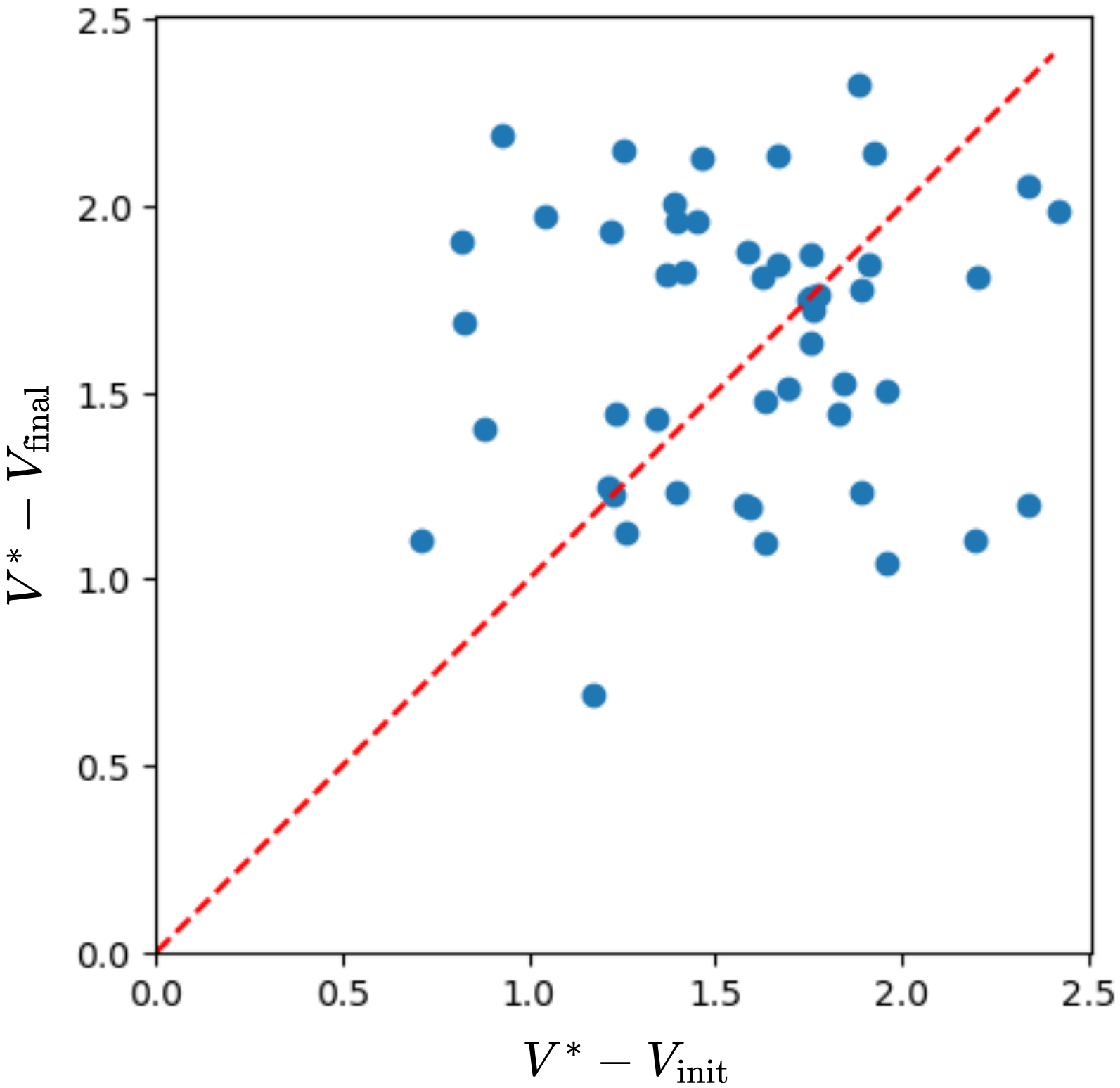}
    \caption{Scatterplot of initial (x-axis) vs. final value-function suboptimality (y-axis) for DQN run on random MDPs (see appendix for the full implementation details). For this plot, we first generated a population of $10$-state, $10$-action MDPs ($100$ in all) by drawing each reward $r(s,a)$ and transition probability $\bP(s'|s,a)$ independently and uniformly from the interval $[0, 1]$ ($\bP$ was also normalized). We then plotted each blue dot by a) picking a random MDP from this population, b) initializing DQN with a random Q-value network, c) finding the difference between $V^*$ and the value of the greedy policy of the initial Q-network (to get the $x$-coordinate of the dot), d) running DQN for a fixed budget of iterations, and e) finding the difference between $V^*$ and the value of the greedy policy of the final Q-network (to get the $y$-coordinate). The red dashed line is the diagonal $y=x$. In our setup, $V_{\text{init}} < V_{\text{final}}$ implies the policy has improved. In that sense, we see that over $50\%$ of the runs lead to a policy worse than the initial, represented by the dots above the diagonal. In fact, on $20\%$ of the runs (the top left corner), the policies significantly degrade.}
    \label{fig:Scatterplot.DQN}
\end{figure}

To get an initial sense of the answers, we present\footnote{The code for generating this plot is at https://tinyurl.com/yuzfvh5y.} Figure~\ref{fig:Scatterplot.DQN}. It shows how the value function of the policies learned by DQN changes across single runs on random MDPs. Clearly, over $50\%$ of the runs result in DQN learning a policy that is worse than the initial guess; on $\approx 20\%$ on the runs, it is in fact significantly worse. While DQN's performance has been extensively studied in specific environments such as Mujoco and Atari, we believe our study is the first to evaluate it over random MDPs. One may view DQN as a complex algorithm and, hence, attribute our observed performance simply to a poor tuning of hyperparameters.
% such as the experience replay length, the target network refresh rate, and the stepsize schedule. 
However, the rest of our work considers linear function approximation to show that these behaviors are consequences of more fundamental issues.
% with DQN's update and sampling rules. 

We focus on the simpler case since answers to our above questions are unknown even for basic Q-learning with linear function approximation and $\epsilon$-greedy exploration. In fact, Problem~1 in `Open Theoretical Questions in Reinforcement Learning'   \cite{sutton1999open}, from 1999, concerns answering similar questions in the context of the closely related linear\footnote{Linear Q-learning (resp. linear SARSA) is  Q-learning (resp. SARSA) with linear function approximation.} SARSA algorithm with $\epsilon$-greedy exploration:

\begin{quote}
    ``... The parameters of the linear function can be shown to have no fixed point in the expected value. Yet neither do they diverge; they seem to `chatter' in the neighborhood of a good policy. This kind of solution can be completely satisfactory in practice, but can it be characterized theoretically? What can be assured about the quality of the chattering solution? New mathematical tools seem necessary."
\end{quote}
Similarly, for linear Q-learning, the following questions have been asked \cite{lu2021convex}: ``Does (it) have a (fixed-point) solution? Does the solution (correspond) to a good policy?" 

\begin{figure*}[t]
\centering 
\hspace{-2.5em}
\begin{subfigure}[b]{0.49\textwidth}
    \centering
    \includegraphics[width=0.9\linewidth
    ]{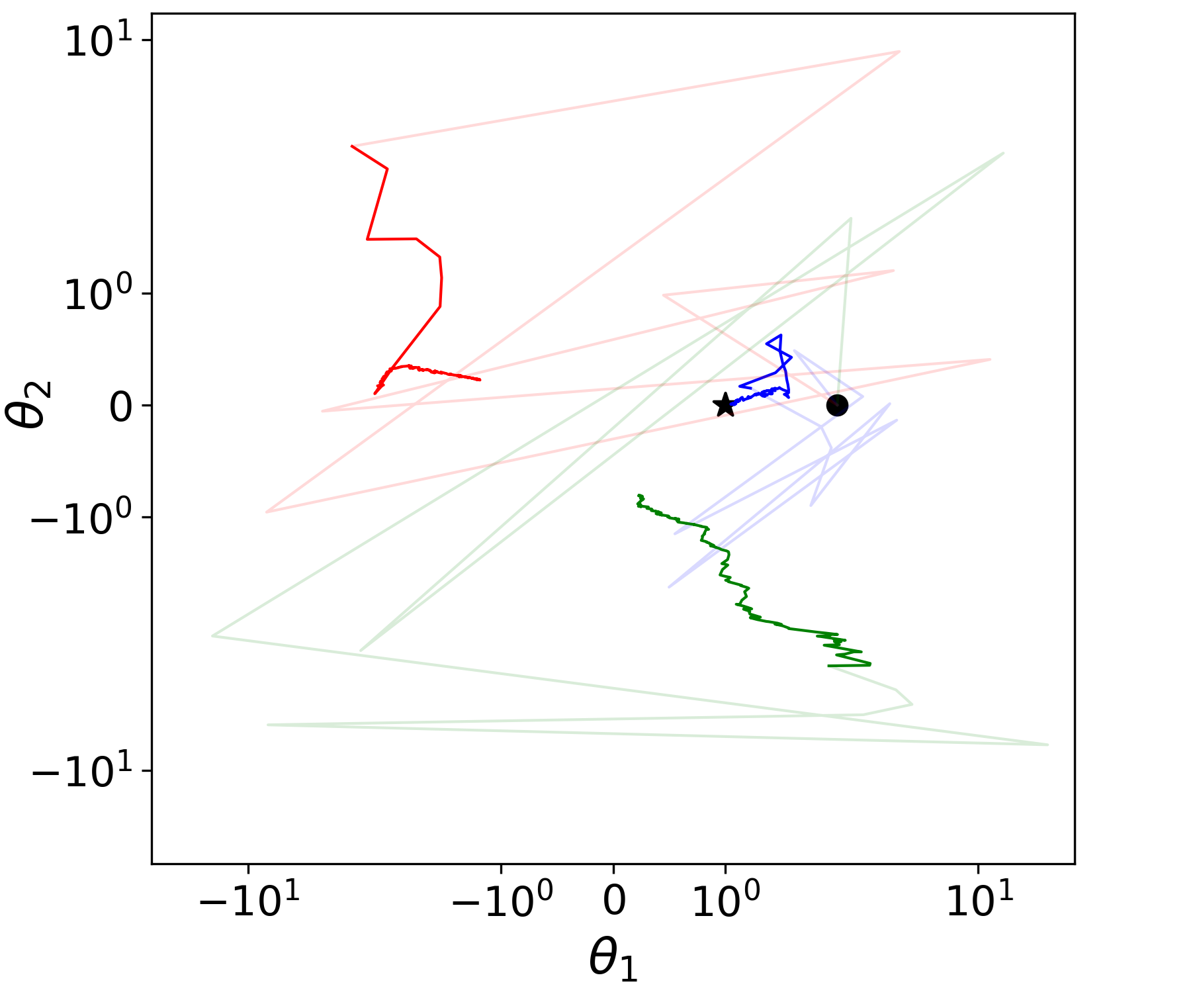}
    \caption{\label{fig:DQN.Trajectories}}
\end{subfigure}
\hspace{-2em}
\begin{minipage}[b]{0.49\textwidth}
    \raisebox{0em}{ % Adjust this value to move the subfigure up or down
    \begin{subfigure}[b]{1\linewidth}
        \centering    
        \includegraphics[width=1.2\linewidth]{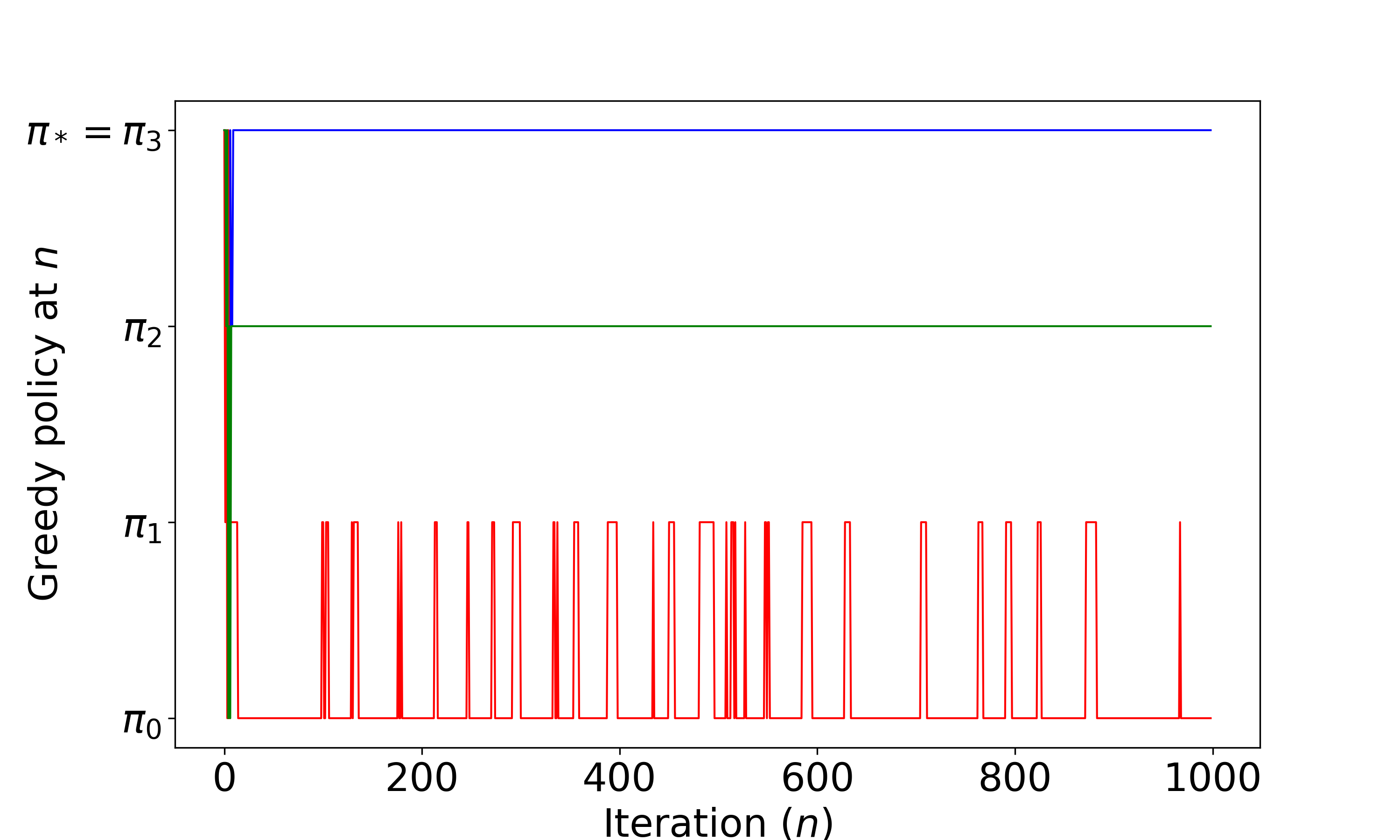}
        \vspace{-0.5em}
        \caption{\label{fig:Chattering}}
    \end{subfigure}
    }
\end{minipage}%
\caption{\label{fig:Linear.DQN} Trajectories of three runs of DQN on a 2-state 2-action MDP with a linear 2-dimensional Q-value approximation which {\em perfectly represents} 
%the optimal Q-value function
$Q^*$ (see the appendix for full implementation details). Figure~\ref{fig:DQN.Trajectories} shows these trajectories in the parameter space (the faded part is the initial behavior). The black star at (1, 0) is $Q^*$'s parameters. All trajectories start at the same place (the black dot), chosen so that the initial behavior is the $\epsilon$-greedy version of $\piS.$ 
Figure~\ref{fig:Chattering} shows the greedy policies associated with the different trajectories.}
\vspace{-2ex}
\end{figure*}

In the remainder of this work, we present empirical and theoretical evidence demonstrating that, in general, the answer to all three of our questions about DQN is \textit{negative} when using linear function approximation. As the first evidence,  look at Figure~\ref{fig:Linear.DQN}, which shows three runs of `linear DQN.' As in a standard DQN \cite{mnih2015human}, this variant also employs $\epsilon$-greedy exploration, experience replay, and a target network. However, it uses a linear function instead of a neural-network-induced nonlinear function for approximating $\QS.$ The reduction in the approximation power is offset by including\footnote{This is ensured by setting one column of the state-action feature matrix to the optimal value function.}  $\QS$ 
in this linear function class. The starting conditions for all three runs are the same, ensuring that the initial policy is close to $\piS.$ In this idealized setting, one would expect linear DQN to always find $\piS.$ Surprisingly, we see three different behaviors: (i) convergence to a sub-optimal policy ({green}), (ii) oscillation between two sub-optimal policies ({red}, tail end) and (iii) convergence to $\piS$ ({blue}). This example already illustrates the \textit{unreliability} of DQN. It also uncovers serious issues beyond instability (divergence to infinity) which a practitioner cannot avoid by just throwing in more data and computation time.

Existing ideas to study Q-learning or SARSA with function approximation are mainly based on the Ordinary Differential Equation (ODE) method. However, these are of limited utility for explaining the above phenomena. To see why, note that the ODE method applies to algorithms of the form 
\begin{equation}
    \label{e:SA.update.rule}
    \theta_{n + 1} = \theta_n + \alpha_n[f(\theta_n) + \rho_n +  M_{n + 1}], \qquad n \geq 0,
\end{equation}
where $f:\bR^d \mapsto \bR^d$ is  some driving function, $\alpha_n$ is a decaying stepsize, $\rho_n$ is some bias term, and $M_{n + 1}$ is the noise.  When $f$ is `nice' overall, e.g., globally Lipschitz continuous, the ODE method can be used to show that the limiting dynamics of \eqref{e:SA.update.rule} is governed by the ODE $\dot{\theta}(t) = f(\theta(t))$  \cite{benaim1999dynamics, borkar2022stochastic}. This niceness holds in \textit{ policy evaluation}. For Q-learning or SARSA, though, $f$ is quite complex: even with linear function approximation, the update rule is nonlinear and involves sampling from distributions that change with $\theta_n.$ So far, ODE-based analyses have succeeded only when these methods are cast as general nonlinear schemes and the state–action distribution induced by the behavior policy is tightly controlled---e.g., via a fixed-policy \cite{carvalho2020new}, a near-optimal policy \cite{melo2008analysis, chen2022finite}, or a smooth soft-max policy \cite{zou2019finite}. With \emph{$\epsilon$-greedy exploration}, the situation is worse since $f$ then is \emph{discontinuous} and \textit{no analysis exists for it}. Moreover, as we discuss in Section~\ref{s:Numerical.Examples}, this discontinuity can introduce new limiting behaviors, e.g., sliding mode, which cannot be explained by continuous ODEs. 

\textbf{Key contributions}: We use Differential Inclusions (DIs) \cite{aubin2012differential}---a set-valued generalization of ODEs---to explain the behaviors of linear Q-learning and linear SARSA with $\epsilon$-greedy exploration (see Section~\ref{s:DI.Primer} for a primer on DIs). Our work's highlights are as follows.
\begin{enumerate}
    \item \emph{Novel analysis framework}: We introduce a new framework (see Section~\ref{s:Setup.Main.Result}) utilizing DI theory to analyze Q-learning and SARSA. Its key steps are: (i) breaking down the parameter space into regions where the algorithm's dynamics are \emph{simple}, (ii) identifying a DI that \emph{stitches} the local dynamics together, and (iii) using this DI to explain the algorithm's overall (possibly complex) behavior. Note that a DI is an extension of an ODE that enables the above stitching by allowing for multiple update directions at every point. 
 
    \item \emph{Explanation of linear Q-learning and linear SARSA(0) behaviors}: Our main result (Theorem~\ref{thm:main.result}) shows that the DIs uncovered by our framework govern \textit{all} asymptotic behaviors of linear Q-learning and linear SARSA(0) employing $\epsilon$-greedy exploration, (idealized) experience replay, and a target network\footnote{Whereas the linear Q-learning variant we analyze theoretically employs idealized experience replay (see Remark~\ref{rem:ind.Sampling} for details), its behavior closely mirrors that of the linear DQN variant introduced earlier. Consequently, we will refer to both variants interchangeably as ``linear DQN," with the context indicating which version is under discussion.}. In Corollary~\ref{cor:fixed.points.proj.Bellman}, we then show that every limit point of linear Q-learning is a projected Bellman fixed point  almost surely (a.s.). With a fixed behavior policy, the projection operator uniquely depends on the behavior policy choice. However, with $\epsilon$-greedy exploration, we show that the projection operator varies depending on the greedy policies associated with the limit point. In this way, we answer the question posed by Sutton in  \cite{sutton1999open} and also show our framework's prowess in explaining the behaviors of linear-DQN-type methods such as those seen in Figure~\ref{fig:Linear.DQN}. 
    
    \item \emph{Discovery of traps that impede learning}: In Section~\ref{s:Numerical.Examples}, we empirically show that linear DQN can be drawn towards projected Bellman fixed points whose induced greedy policies are suboptimal. Such attractors act as traps that prevent the algorithm from learning better policies. In fact, we identify an MDP setting where linear DQN a.s. identifies the worst possible policy. This example concretely shows that the answer to each of our three questions is negative in the context of linear DQN. Finally, we remark that while our analysis also applies to tabular Q-learning, in that case a global Lyapunov function ensures that no local traps can exist.
\end{enumerate}

\textbf{Related work}: Several works report various pathological behaviors for approximate value-function-based methods. In planning, \cite{bertsekas1996neuro} argues that approximate policy iteration may generally be prone to policy oscillations, chattering, and convergence to poor solutions. Similarly, \cite{de2000existence} shows how approximate value iteration may oscillate forever and not possess any fixed points. Within RL, 
\cite{gordon1996chattering, gordon2000reinforcement} and \cite{zhang2023convergence} discuss chattering in linear SARSA(0), but formally show only convergence to a bounded region. More recently, \cite{young2020understanding}, \cite{schaul2022phenomenon}, and \cite{patterson2024empirical} empirically discuss the above pathological behaviors in approximate value-function-based RL methods with greedification. Our work is the first to rigorously explain these phenomena in RL.

Within the Q-learning literature, a prominent stream uses the ODE method  
to analyze the linear \cite{melo2008analysis, carvalho2020new,chen2022finite} and nonlinear (neural) function approximation \cite{fan2020theoretical,xu2020finite} variants. However, these works {\em hold the behavior policy fixed} and impose other conditions such as this policy
%n it (e.g., 
being close to the optimal policy. These assumptions 
ensure that the resulting nonlinear ODE %nonlinear SA 
has a Lyapunov function and thus convergence guarantees. Another such notable work is \cite{lee2020unified}, which uses the switched system theory for analysis. None of these analyses carry over to the $\epsilon$-greedy exploration case because the behavior policy and the resultant dynamics discontinuously change.

There are also analyses that apply ODE methods 
to study SARSA(0) with changing policies  \cite{ melo2008analysis, zou2019finite}. However, these apply only when the policy improvement operator is Lipschitz continuous with a sufficiently small Lipschitz constant, which ensures the limiting ODE is `very smooth.' This restrictive condition holds, e.g., for softmax-type policies with a sufficiently small inverse-temperature parameter. Hence, these analyses reveal very little about the behavior under \textit{discontinuous} $\epsilon$-greedy exploration (the case when the inverse temperature parameter is $\infty$). 

A few variants of  Q-learning have already been analyzed using DI-based approaches \cite{maei2010toward, bhatnagar2016multiscale, avrachenkov2021full}. However, they use DIs for other reasons: the use of sub-gradients, or an intrinsic problem having multiple solutions. This is fundamentally different from our need, which stems from the discontinuity of $\epsilon$-greedy exploration. Finally, \cite{wunder2010classes} and \cite{banchio2023adaptive} use DIs to shed light on the dynamics of (tabular) Q-learning in stateless, multi-agent repeated games.

\section{Preliminaries}
This section has two distinct parts: the first provides a brief background on Q-learning and SARSA with linear function approximation and $\epsilon$-greedy exploration; the second gives a concise introduction to DIs. 

\subsection{Linear Q-learning and SARSA with $\epsilon$-greedy Policy}
\label{ss:algo}
For a set $U,$ let $\chi(U)$ denote the set of probability measures on it. Our setup is that of an MDP $(\cS, \cA, \gamma, \bP, r),$ where $\cS$ is a finite state space, $\cA$ is a finite action space equipped with a total order, $\gamma \in [0, 1)$ is the discount factor, and $\bP: \cS \times \cA \to \chi(\cS)$ and $r: \cS \times \cA \times \cS \to \bR$ are functions such that $\bP(s, a)(s') \equiv \bP(s'|s, a)$ specifies the probability of moving from a state $s$ to $s'$ under action $a,$ while $r(s, a, s')$ is the one-step reward obtained in this transition. Let $\QS \in \bR^{|\cS||\cA|}$ be the optimal Q value function of this MDP, and $\Phi \in \bR^{|\cS||\cA| \times d}$ the given feature matrix. Our goal, aligned with standard RL, is to find a $\thS \in \bR^d$ such that $\QS \approx \Phi \thS.$ 

Two algorithms to find such a $\thS$ are linear Q-learning and linear SARSA(0) with $\epsilon$-greedy exploration. Various forms of these algorithms have been discussed in the literature, ranging from the plain vanilla type to more sophisticated ones with a replay buffer and a target network. Up to some idealization, all these variants can be expressed via a single template update rule which we now describe.

Let $\epsilon \in [0, 1)$ be the greedy-exploration parameter and $\epsilon' \in [0, 1]$ the action-sampling parameter at the succeeding state. Further let $\ell \geq 0$ and $\mu \equiv (\mu_0, \ldots, \mu_{\ell})$ be the replay-buffer length and an associated buffer-sampling distribution. Also, let $\Delta \in (0, 1]$ be the rate at which the target-network estimate is updated. Finally, let $\thetam_{0}, \theta_0, \ldots,\theta_{-\ell} \in \bR^d$ be some initial estimates of $\thS$. Then, for $n \geq 0,$ an unified update rule for linear Q-learning and linear SARSA(0) is
\begin{equation}
\label{e:generic.update} 
    \theta_{n + 1} = {} \theta_n + \alpha_n \delta_n \phi(s_n, a_n),
\end{equation}
where
\begin{equation}
\label{e:deltan.defn}
    \delta_n = {} r(s_n, a_n, s'_n) + \gamma \phi^\tr(s'_{n}, a_n')\thetam_{n} - \phi^\tr(s_n, a_n) \theta_n.
\end{equation}
In this update rule, $\alpha_n \in \bR_{\geq 0}$ is the stepsize, $\theta_n \in \bR^d$  is the current estimate of $\thS,$ while $\thetam_n \in \bR^d,$ $n \geq 1,$ is the output of the target network. Further, $(\thetam_n)_{n \geq 0}$ is updated\footnote{In practice, the target network is updated after every $1/\Delta$-many steps for some $\Delta \in (0, 1)$. We idealize this by presuming that the target-network estimate is updated with probability $\Delta$ in every step.} using
\begin{equation}
    \label{e:target.network.update}
    \thetam_{n + 1} = \thetam_n + \tau_n (\theta_n - \thetam_n) \zeta_{n + 1},
\end{equation}
where $(\zeta_{n})$ is a sequence of Independent and Identically Distributed (IID) Bernoulli random variables with mean $\Delta,$ and $(\tau_n)$ is another stepsize sequence. Next, $\phi^\tr(s, a),$ with $^\tr$ being transpose, denotes the $(s, a)$-th row of $\Phi,$ while  $\delta_n$ is the one-step Temporal-Difference (TD) error. The next paragraph describes how  $(s_n, a_n)$ and $(s_n', a_n')$ are sampled. 

Let $\pie_n: \cS \to \chi(\cA)$ be the $\epsilon$-greedy policy at time $n \geq -\ell,$ i.e., the policy that samples the greedy action w.r.t. $\Phi \theta_n$ (the current $\QS$ estimate) with probability $1 - \epsilon$ and a random action with probability $\epsilon.$ In mathematical notations,
\begin{equation}
\label{e:eps.greedy.policy}
    \pie_n(a|s) = 
    \begin{cases}
    1 - \epsilon  + \dfrac{\epsilon}{|\cA|},
    & a = \underset{a'}{\arg \max} \, \phi^\tr (s, a') \theta_n, \\[1ex]
    \dfrac{\epsilon}{|\cA|}, & \text{ otherwise.}
    \end{cases}
\end{equation}
We presume that $\arg \max$ breaks ties using the total order on $\cA.$ Similarly, define $\piep_n$ with respect to $\thetam_n.$ Noting that the state and action spaces are finite, the number of $\epsilon$-greedy policies is finite. We suppose throughout that these policies satisfy the following condition. %
{
\renewcommand{\theenumi}{$\cB_\arabic{enumi}$}
\begin{enumerate}
    \setcounter{enumi}{0}
    
    \item \label{a:ergodicity} The Markov chain induced by each $\epsilon$-greedy policy is ergodic or, equivalently, aperiodic and irreducible (and hence has a unique stationary distribution).
\end{enumerate}
}
\noindent For $n \geq -\ell,$ let $\de_n$ be the stationary distribution associated with the Markov chain\footnote{At any time $t \geq 0,$ this Markov chain moves from state $s$ to $s'$ with probability $\sum_{a}\pie_n(a|s) \bP(s'|s, a).$} 
induced by $\pie_n.$ Then, for each $n \geq 0,$ $(s_n, a_n)$ and $(s'_n, a_n')$ are sampled as follows. First, an  index $k \in \{0, \ldots, \ell\}$ is sampled from $\mu;$ then, $s_n$ is sampled from $\de_{n - k}$ and $a_n$ from $\pie_{n - k}(\cdot|s_n);$ finally, $s_n'$ is sampled from $\bP(\cdot| s_n, a_n)$ and $a_n'$ from $\piep_n(\cdot|s_n').$ These five samples are drawn with independent randomness.

\begin{remark}
\label{rem:eps'.choice}
Note that \eqref{e:generic.update} with $\epsilon' = 0$ (resp. $\epsilon' = \epsilon$) is linear Q-learning (resp. SARSA(0)) with $\epsilon$-greedy exploration. Specifically, the $\max$ operator with which Q-learning is usually written is implicitly specified via the manner in which action $a_n'$ is sampled (from the $\epsilon'$-greedy policy). 
\end{remark}

\begin{remark}
\label{rem:ind.Sampling}
The sampling choice for $s_n$ leads to different variants of Q-learning and SARSA. In standard DQN, $s_n$ is randomly sampled from a replay buffer that holds a sufficiently long but finite record of all the states observed recently. Our way of sampling $s_n$ from the stationary distributions associated with $\theta_n, \ldots, \theta_{n - \ell}$ serves as an idealized version of that strategy. In Section~\ref{s:Numerical.Examples}, we show that our idealized algorithm has similar behaviors as in Figure~\ref{fig:Linear.DQN}. 
\end{remark}

\subsection{Primer on Differential Inclusions}
\label{s:DI.Primer}
    A DI is a relation of the form $\dot{\theta}(t) \in h(\theta(t))$ where $h(\theta)$ is a non-empty \textit{subset} of $\bR^d$ for each $\theta \in \bR^d.$ Its solution is any absolutely continuous function $t \mapsto \theta(t)$ that satisfies the given DI relation \textit{for almost all $t \in \bR$} and an initial condition like  $\theta(0) = \theta_0.$ A DI's solution, though, need not be unique. Unlike an ODE $\dot{\theta}(t) = f(\theta(t))$, which only allows for a single velocity $f(\theta)$ at each point $\theta,$ the DI $\dot{\theta}(t) \in h(\theta(t))$ admits an entire set. It’s as if, at each $t,$ the solution trajectory can pick any arrow inside a little ``cloud" of allowed velocities. As we show, this extra ``wiggle room"  makes DIs ideal for capturing mode switches that are inherent in linear DQN-type methods.
    
    To see an illustration (cf. \cite[(11)]{cortes2008discontinuous}) of how DIs can help handle multiple modes, consider \eqref{e:SA.update.rule} with $d = 1,$ $f(\theta) = -1$ (resp. +1) for $\theta > 0$ (resp. $\theta \leq 0$), and no bias or noise (i.e., $\rho_n, M_{n + 1} \equiv 0$).  Clearly, this algorithm operates differently along the positive and the negative real line. Due to decaying stepsizes, its iterates should converge to $0.$ However, this behavior cannot be studied via the ODE $\dot{\theta}(t) = f(\theta(t))$ for which the origin is not even an equilibrium point. In fact, this ODE has no solution at $0$: there exists no $t \mapsto \theta(t)$ map with $\theta(0) = 0$ and $\dot{\theta}(t) = f(\theta(t));$ the natural choices: $\theta(t) = -t,$ $\theta(t) = +t,$ or $\theta(t) \equiv 0$ do not satisfy the ODE relation.

    The dynamics of the above algorithm, though, can be studied using the DI $\dot{\theta}(t) \in h(\theta(t)),$ where $h(\theta) = \{+1\}$ (resp. $\{-1\}$) when $\theta < 0$ (resp. $\theta > 0$), and equals the interval $[-1, + 1]$ for  $\theta = 0.$ Since $h(0)$ is the convex closure of the set $\{-1, + 1\},$ it can be shown that $h$ is \textit{Marchaud}, i.e.,  \textit{Lipschitz continuous in a set-valued sense} (see $\cC_1$ in  Theorem~\ref{thm:SRI.convergence} for details). Like Lipschitz continuity guarantees the existence of solutions for ODEs (for any initial point), the Marchaud property does so for a DI. Thus, the above DI has a solution for any initial condition: even for $\theta(0) = 0,$ $\theta(t) \equiv 0$ now is a valid solution. In fact, the origin can be shown to be a unique global attractor. This last fact can be used to establish the convergence of the above algorithm to the origin.

\section{Key contributions: Our analysis framework \& its application to linear Q-learning/SARSA}
\label{s:Setup.Main.Result}
This section has two subsections. The first introduces our analysis framework, while the second uses it to systematically explain all asymptotic behaviors of linear Q-learning and linear SARSA(0) with $\epsilon$-greedy exploration. The proofs for all the results stated here are given in Section~\ref{s:Proofs}. 

\subsection{Our Analysis Framework}
We propose the following three steps to analyze an update rule like \eqref{e:SA.update.rule} when $f$ is discontinuous. The novelty, specifically in Steps 1) and 3), lies in its application within RL. 
\begin{enumerate}
    \item \emph{Partition the parameter space $\bR^d$ into regions over which $f$ is `simple'}: What counts as ``simple" will depend on the algorithm. For \eqref{e:generic.update}, i.e., linear Q-learning and linear SARSA with $\epsilon$-greedy exploration, we partition $\bR^d$ such that each subset is made up of those $\theta$'s where the $\epsilon$-greedy policy is the same; see \eqref{e:Ra.defn} for details. Lemma~\ref{lem:base.driving.function} shows that the $f$ associated with \eqref{e:generic.update} is \emph{linear} and \textit{continuous} within these greedy regions, but changes \emph{discontinuously} across region boundaries.     
    
    \item \emph{Use `Filippov\footnote{Although “Krasovskii convexification” is the more precise term, we follow common practice and refer to it as the “Filippov convexification.”} convexification' to stitch the different $f$-definitions and make a DI}: Formally, this stitching is to be done via the set-valued map $h: \bR^d \to 2^{\bR^d}$ (power set of $\bR^d$) given by 
    \begin{equation}
    \label{e:filippov.stitch}
        h(\theta) = \bigcap_{\delta > 0} \overline{\co} (f(B(\theta, \delta))).
    \end{equation}
    Here, $\overline{\co}$ is the convex closure. Further, $B(\theta, \delta)$ and $f(B(\theta, \delta))$ mean the open ball of radius $\delta$ at $\theta,$ and  its image under $f,$ respectively. The set $h(\theta)$ is the singleton set $\{f(\theta)\}$ if $f$ is continuous at $\theta,$ and all convex combinations of $f$ values otherwise.
    The DI to study \eqref{e:SA.update.rule}'s overall behavior is %\cite{aubin2012differential}
    \begin{equation}
    \label{e:lim.DI}
        \dot{\theta}(t) \in h(\theta(t)).
    \end{equation}
    Clearly, the DI for Section~\ref{s:DI.Primer}'s example matches the one obtained via the above construction. We remark that Filippov construction is commonly employed in control theory to deal with discontinuous dynamics \cite{filippov2013differential}. 
    
    \item \emph{Establish a formal link between the DI and the algorithm's dynamics}: The aim here is to show that the discrete-time iterates  $(\theta_n)$ of  \eqref{e:SA.update.rule} eventually track a solution of the (non-stochastic) DI in \eqref{e:lim.DI}. To prove this claim, one typically has to show that $h$ is Marchaud, i.e., continuous in a set-valued sense, and that the stepsizes decay sufficiently fast so that the cumulative noise and bias effect is negligible. The asymptotics of the DI solutions can then be used to explain all limiting behaviors of \eqref{e:SA.update.rule}. In our work, we build upon \cite{borkar2022stochastic} to rigorously establish all the above properties. The novelty in our proof lies in how we handle the bias effect.
\end{enumerate} 

\subsection{Analysis of Linear Q-learning/SARSA with $\epsilon$-greedy}
\label{s:appl}
We now use our framework to analyze the limiting behaviors of linear Q-learning and linear SARSA(0) with $\epsilon$-greedy exploration, (idealized) experience replay, and a target network. 

\subsubsection{Analysis Step~1 (partitioning $\bR^d$)}
\label{s:Step_1}
We first show how  \eqref{e:generic.update}
can be rewritten as \eqref{e:SA.update.rule}. For any $\thetam, \theta^{(0)}, \ldots, \theta^{(-\ell)} \in \bR^d,$ let 
\begin{equation}
\label{e:v.defn}
    v(\thetam, \theta^{(0)}, \ldots, \theta^{(-\ell)}) 
    :=
    \bE\bigg[\delta_0 \phi(s_0, a_0) \bigg| \thetam_0 = \thetam,     \theta_{k} = \theta^{(k)}, k = -\ell, \ldots, 0 \bigg].
\end{equation}
The above expectation is with respect to $s_0, a_0,s_0',$ and $a_0'.$ These random variables are sampled as per the procedure specified below Assumption~\ref{a:ergodicity}. That is, an  index $k \in \{0, \ldots, \ell\}$ is first sampled from $\mu,$ the buffer-sampling distribution. Then, $s_0$ is sampled from $\de_{- k},$ the stationary distribution of the Markov chain induced by the $\epsilon$-greedy  policy $\pie_{- k}$ defined with respect to $\Phi \theta^{(-k)}.$ Next, $a_0$ is sampled from $\pie_{- k}(\cdot|s_0)$ and $s_0'$ is sampled from $\bP(\cdot| s_0, a_0).$ Finally, $a_0'$ is sampled from $\piep_{0}(\cdot|s_0'),$ the $\epsilon'$-greedy policy defined with respect to $\Phi \thetam_0.$

Next, for any $\theta \in \bR^d,$ let 
\begin{equation}
\label{e:f.Defn}
    f(\theta) := v(\theta, \theta, \ldots, \theta),
\end{equation}
and, for $n \geq 0,$ let 
\begin{align}
    \rho_n := {} & v(\thetam_n, \theta_n, \ldots, \theta_{n - \ell}) - f(\theta_n) \label{e:rhon.defn}\\
    \intertext{and} 
    M_{n + 1} := {} & \delta_n \phi(s_n, a_n) - v(\thetam_n, \theta_n, \ldots, \theta_{n - \ell}) \label{e:Mn.defn}.
\end{align}
Using the above definitions, it is easy to see that \eqref{e:generic.update} can be expressed in the form given in \eqref{e:SA.update.rule}. 

Next, we describe the way we partition  $\bR^d.$ Let $\cA^{\cS} :=\{\ba: \cS \to \cA\}$ denote the set of all possible deterministic policies. For a policy $\ba \in \cA^{\cS},$ let %the associated \textit{greedy region} 
\begin{equation}
\label{e:Ra.defn}
    \cR_{\ba} := \{\theta \in \bR^d: \forall s \in \cS, \ba(s) = \arg\max_a \phi^\tr(s, a) \theta\},
\end{equation}
where we break ties in $\arg \max$ using the total order. Clearly, for any $\theta \in \bR^d,$ there is a unique $\ba$ such that $\theta \in \cR_\ba.$ Thus, $\{\cR_\ba : \ba \in \cA^{\cS}\}$ partitions $\bR^d,$ and this is the one we use throughout this work.
% It is possible that $\cR_\ba = \emptyset$ for some $\ba.$ 
For $\ba$ where $\cR_\ba \neq \emptyset,$ the greedy (hence, $\epsilon$-greedy) policy corresponding to $\Phi \theta$ is the same for every $\theta \in \cR_\ba,$ and it is $\ba.$ Hence, we refer to $\cR_{\ba}$ as the \textit{greedy region} associated to $\ba.$ Finally, note that each $\cR_{\ba}$ is a cone since $\theta \in \cR_\ba \implies c\theta \in \cR_\ba$ for any scalar $c > 0.$ 

The advantage of the above partition is that $f$ has a simple linear form in each region, which we now describe. We need a few notations to state this result. Let $\pie_\ba$ (resp. $\piep_\ba$)
be the $\epsilon$-randomization (resp. $\epsilon'$-randomization) of the policy $\ba.$ That is, at any state $s,$ $\pie_\ba$ picks a random action with probability $\epsilon$ and $\ba(s),$ the action prescribed by $\ba,$ with probability $1 - \epsilon.$ Clearly, $\pie_n = \pie_\ba$ (resp. $\piep_n = \piep_\ba$) whenever $\theta_n \in \cR_\ba$ (resp. $\thetam_n \in \cR_\ba$). Next, let $\de_\ba$ denote the stationary distribution associated with the Markov chain induced by $\pie_\ba,$ and let 
\begin{equation}
\label{e:b_ba.defn}
    b_{\ba} := \bE[\phi(s, a) r(s, a, s')] =  \Phi^\tr \De_{\ba} \br
\end{equation}
and
\begin{align}
    A_\ba := {} & \bE[\phi(s, a) \phi^\tr(s, a) - \gamma \phi(s, a) \phi^\tr(s', a')] \nonumber\\
    = {} & \Phi^\tr \De_{\ba} (\bI - \gamma \Pep_\ba) \Phi. \label{e:A_ba.defn}
\end{align}
In the above definitions, the expectation is with respect to $s \sim \de_\ba,$ $a \sim \pie_\ba (\cdot|s),$ $s' \sim \bP(\cdot|s, a),$ and $a' \sim \piep_\ba(\cdot|s').$ Further, $\De_\ba$ is the diagonal matrix of size $|\cS||\cA| \times |\cS||\cA|$  whose $(s, a)$-th diagonal entry is $\de_\ba(s) \pie_\ba(a|s),$ $\br$ is the $|\cS||\cA|$-dimensional vector whose $(s, a)$-th coordinate is $\br(s, a) = \sum_{s' \in \cS} \bP(s'|s, a) r(s, a, s'),$  while $\Pep_\ba$ is the matrix of size $|\cS||\cA| \times |\cS||\cA|$ with $\Pep_\ba((s, a), (s', a')) = \bP(s'|s, a) \piep_\ba(a'| s').$ 

% Then, when $\theta_n \in \cR_{\ba},$ the expectations of the different terms in $\delta_n \phi(s_n, a_n)$ conditional on $\cF_n$ are

\begin{lemma}
\label{lem:base.driving.function}
For any $\theta \in \bR^d,$ the function $f$ given in \eqref{e:f.Defn} satisfies $f(\theta) = \sum_{\ba \in \cA^{\cS}}  \left(b_{\ba} - A_{\ba} \theta\right) \indc[\theta \in \cR_{\ba}].$
\end{lemma}

\begin{remark}
While $f$ is nonlinear overall, Lemma~\ref{lem:base.driving.function} shows that it is piece-wise linear. That is, $f(\theta) = b_\ba - A_\ba \theta$ for $\theta \in \cR_\ba,$ and this definition changes discontinuously from one greedy region to the other. For $\epsilon = \epsilon',$ $f|_{\cR_\ba}$ is the driving function that governs the behavior of TD(0) with linear function approximation for evaluating the policy $\pie_{\ba}$ \cite[(9.11)]{sutton2018reinforcement}.   Figure~\ref{fig:vector_field} shows the partitions and the nature of $f$ over each sub-region for two different MDP settings. 
\end{remark}

\subsubsection{Analysis Step 2 (DI identification)} 
\label{s:Step_2}
The DI to study the limiting dynamics of \eqref{e:generic.update} is the one given in \eqref{e:lim.DI}, where the set-valued map $h: \bR^d \to 2^{\bR^d}$ from \eqref{e:filippov.stitch} is defined using the function $f$ from \eqref{e:f.Defn} (or, equivalently, the one in Lemma~\ref{lem:base.driving.function}). Henceforth, we refer to this DI as the limiting DI of \eqref{e:generic.update}.

In  Lemma~\ref{lem:htheta} below, we give an equivalent but simpler description of this specific function $h.$ For $\theta \in \bR^d,$ let 
\[
    \supp(\theta) = \left\{\ba \in \cA^\cS: \phi^\tr (s, \ba(s)) \theta = \max_{a \in \cA} \phi^\tr(s, a) \theta\ \forall s \in \cS \right\}.
\]
Clearly, $1 \leq |\supp(\theta)| \leq |\cA|^{|\cS|}$ since $\{\cR_\ba\}$ partitions $\bR^d.$ In particular, if $\theta$ is in the interior of $\cR_\ba$ for some $\ba,$  then $\supp(\theta) = \{\ba\};$ for the one on the boundary, $|\supp(\theta)| \geq 2.$

\begin{lemma}
\label{lem:htheta}
For any $\theta \in \bR^d,$ the function $h$ in the limiting DI of \eqref{e:generic.update} satisfies $h(\theta) = \co\left\{ b_\ba - A_\ba \theta: \ba \in \supp(\theta) \right\},$ 
where $\co$ is the convex hull. Specifically,  $h(\theta) = \{b_\ba - A_\ba \theta\}$ for any $\theta$ in the interior of $\cR_\ba.$ Further,  $f(\theta) \in h(\theta).$ 
\end{lemma}

\subsubsection{Analysis Step 3 (algorithm-DI connection)} 

Our main result (Theorem~\ref{thm:main.result}) is that \eqref{e:generic.update}'s limiting DI completely governs its limiting dynamics. To state this result, we need two additional assumptions. Let $\|\cdot\|$ denote the Euclidean norm.
{
\renewcommand{\theenumi}{$\cB_\arabic{enumi}$}
\begin{enumerate}
    \setcounter{enumi}{1}
%
    % \item \label{a:reward} There exists a constant $K_r \geq 0$ such that $|r(s, a, s')| \leq K_r$ $\forall s, s' \in \cS$ and $a \in \cA.$
    
    \item \label{a:stepsize}  $(\alpha_n)$ satisfies $\sup_{n \geq 0} \alpha_n \leq 1,$ $\sum_{n \geq 0} \alpha_n = \infty$ and $\sum_{n \geq 0} \alpha_n^2 < \infty.$ Similarly, $(\tau_n)$ satisfies $\sup_{n \geq 0} \tau_n \leq 1,$ $\sum_{n \geq 0} \tau_n  = \infty,$ $\sum_{n \geq 0} \tau_n^2 < \infty,$ and $\alpha_n/\tau_n \to 0.$

    \vspace{0.5ex}
    
    \item \label{a:phi} $\Phi$ has full column rank.
    
    % Also, $\exists K_\phi  \geq 0$ such that $\|\phi(s, a)\| \leq K_\phi$ $\forall s, s' \in \cS$ and $a \in \cA.$

    % \item \label{a:PD.matrix} For all $\ba \in \cA^\cS,$ $A_\ba$ is positive definite, i.e., $\theta^\tr A_\ba \theta > 0 \, \forall \theta \in \bR^d$ ($A_\ba$ need not be symmetric). 
%
\end{enumerate}
}

We also need a few definitions. In relation to \eqref{e:lim.DI}, we will say a set $\Gamma \subseteq \bR^d$ is \textit{invariant} if, for every $\theta_0 \in \Gamma,$ there is \textit{some} solution trajectory $(\theta(t))_{t \in (-\infty, \infty)}$ of \eqref{e:lim.DI} with $\theta(0) = \theta_0$ that lies entirely in $\Gamma.$ %It is important to note that this condition is not required to be satisfied by every solution of \eqref{e:app.DI} satisfying $\theta(0) = x.$ 
An invariant set $\Gamma$ is additionally \textit{internally chain transitive} if it is compact and satisfies the following property: for $x, y \in \Gamma,$ $\nu > 0,$ and $T > 0,$ there exist $m \geq 1$ and points $x_0 = x, x_1, \ldots, x_{m - 1}, x_m = y$ in $\Gamma$ such that a solution trajectory of \eqref{e:lim.DI} initiated at $x_i$ meets the $\nu$-neighborhood of $x_{i + 1}$ for $0 \leq i < m$ after a time that is equal or larger than $T.$ Such characterizations are useful to restrict the possible sets to which \eqref{e:generic.update} could converge to. For example, for the DI in Section~\ref{s:DI.Primer}, while $\bR, [0, \infty), (-\infty, 0],$ and $\{0\}$ are all invariant, only $\{0\}$ is internally chain transitive.

\begin{theorem}[\textbf{Main Result}]
\label{thm:main.result}
Suppose \ref{a:ergodicity}, \ref{a:stepsize}, and  \ref{a:phi} hold. Then $(\theta_n)$ obtained from \eqref{e:generic.update} converges to a closed, connected, internally chain transitive invariant set of its limiting DI a.s. on the event $\{\sup_n \|\theta_n\| < \infty\}.$ 
% The limiting DI is the one given in \eqref{e:lim.DI} with $h$ as in \eqref{e:alt.h.Defn}. 
%
\end{theorem}

\begin{remark}
    Our result states that $(\theta_n)$  either diverges to $\infty$ or converges to a suitable (possibly sample-point dependent)  invariant set of its limiting DI. 
    % This limiting DI could have many such invariant sets, in which case, the iterates might converge to different invariant sets on different runs. 
    In this way, our result captures all possible limiting behaviors of \eqref{e:generic.update} and resolves the open question in  \cite[Problem~1]{sutton1999open}. Notably, our result is the first to characterize the asymptotic behaviors of any value-function-based algorithm with function approximation and $\epsilon$-greedy exploration. Our proof's key challenge and novelty is in taming the perturbation term $\rho_n$ that arises due to the use of experience replay and a target network.    
    %
    % In Section~\ref{s:Numerical.Examples}, we show that the convergence to an invariant set does not guarantee the superiority of a resulting limiting greedy policy over intermediate policies. \hfill $\blacksquare$
    %
\end{remark}
    
\begin{remark}
    \label{rem:hyp.ind}
    The limiting DI for \eqref{e:generic.update} does not depend on the hyperparameters such as experience replay length $\ell,$ target-network refresh rate $\Delta,$ and the stepsizes  $(\alpha_n)$ and $(\tau_n).$ This means that adjusting these hyperparameters does not change the possible limiting sets for the sequence $(\theta_n).$  
\end{remark}

\begin{remark}
    \label{rem:stability}
    There are two important scenarios where the iterates obtained using \eqref{e:generic.update} are already known to be a.s. stable, i.e., $\Pr\{\sup_{n \geq 0} \|\theta_n\| < \infty\} = 1.$ In these cases, our claim holds on almost every sample point. The first case is that of linear SARSA(0) with $\epsilon$-greedy exploration ($\epsilon' = \epsilon$), but without experience replay ($\ell = 0$) or a target network ($\thetam_n = \theta_n$). Its stability has been established in \cite{gordon2000reinforcement}. The second case is that of tabular Q-learning ($\Phi = \bI$), whose stability follows using a simple inductive argument, e.g., see \cite{gosavi2006boundedness}. 
\end{remark}

We next show that every potential limit point of linear DQN is a fixed point of a suitable projected Bellman operator. Let $T, T_\ba: \bR^{|\cS||\cA|} \to \bR^{|\cS||\cA|}$ be given by
\[
    TQ(s, a) = \br(s, a) + \gamma \sum_{s'} \bP(s'|s, a) \max_{a'} Q(s', a')
\]
and

\[
    T_\ba Q (s, a)  = \br (s, a) + \gamma \sum_{s', a'} \Pep_\ba(s', a'|s, a) Q(s', a'),
\] 
where $\Pep_\ba$ is as defined below \eqref{e:A_ba.defn}.

 \begin{corollary}
\label{cor:fixed.points.proj.Bellman}
    Suppose \ref{a:ergodicity}, \ref{a:stepsize}, and \ref{a:phi} hold, and that the linear DQN iterates converge to a limit $\thS$. Almost surely then:
\begin{enumerate}
    \item The vector $\thS$ is a zero or an equilibrium point of the limiting DI, i.e., $0 \in h(\thS).$
    
    \item There exists a deterministic policy $\pi_\ba$, greedy with respect to $\Phi\thS$, and a norm $\|\cdot\|$ such that
    \[
        \Pi\,T_\ba \,\Phi\theta^* \;=\; \Pi\,T\,\Phi\theta^* \;=\;\Phi\theta^*,
    \]
    where $\Pi$ is $\|\cdot\|$-projection onto the column space of $\Phi.$
\end{enumerate}
\end{corollary}

\begin{remark}
\label{rem:linear.DQN.limitations}
    Corollary~\ref{cor:fixed.points.proj.Bellman} shows a novel insight: every limit of linear DQN---even when it lies on the boundary between two or more greedy regions---is a fixed point of a projected Bellman operator. Thus, linear DQN does its expected job, but, as Section~\ref{s:Numerical.Examples} shows, the greedy policies associated with such fixed points need not be optimal---or even near-optimal. 

    This last fact can explain all of linear DQN's shortcomings. In fact, Figure~\ref{fig:2PartitionOneSelfConsistent} discusses an MDP example, wherein linear DQN a.s. converges to a global attractor (the blue diamond), but the associated greedy policy is the worst possible. This conclusion is independent of how we choose the stepsize sequences $(\alpha_n)$ and $(\tau_n),$ the target network refresh rate $\Delta,$ the buffer sampling length $\ell,$ or the buffer sampling distribution $\mu_0, \ldots, \mu_\ell.$ Moreover, since the local dynamics in each greedy region is linear, the above conclusion holds even under small perturbations of the underlying MDP parameters.

    Hence, for linear DQN, the answer to all three questions from Section~\ref{s:Introduction} is negative: it offers no guarantee of monotonic policy improvement, no assurance of convergence to a locally optimal policy (howsoever locality is defined), and no promise that the final policy will outperform the initial one. 
\end{remark}

% Thus, the answer to all the three questions we posed is, in general, no in the linear DQN context. 

% This global attractor is also a fixed point of a projected Belm

% In this case, the linear DQN iterates will almost surely con

% Thus, all solution trajectories of the limiting DI will converge to this attractor. 

% Separately, for this example' MDP, there

% zero. Theorem~\ref{thm:main.result} (our main result) then implies that the stochastic linear DQN iterates will almost surely also converge to this zero. Hence, the greedy policy associated with the asymptotic linear DQN iterates will almost surely be the worst possible one. 

%      Unfortunately, as Figure~\ref{fig:vector_field} illustrates, these fixed points typically lie within greedy regions tied to sub‑optimal policies and, thereby, act as traps that halt further learning.

% \end{remark}

Lastly, we discuss the convergence of tabular Q-learning with $\epsilon$-greedy exploration, idealized experience replay, and a target network. Theorem~\ref{thm:main.result}, along with Remark~\ref{rem:stability}, shows that the iterates must a.s. converge to some invariant set of its limiting DI. Our next result establishes that this invariant set must necessarily be the singleton set $\{\QS\},$ as expected.

\begin{proposition}[Tabular Q-learning with $\epsilon$-greedy Policy]
\label{prop:tabular}
Suppose \ref{a:ergodicity}, \ref{a:stepsize} hold, and \eqref{e:generic.update} corresponds to tabular Q-learning with $\epsilon$-greedy exploration, i.e., suppose $\epsilon' = 0$ and $\Phi = \bI.$ Then, $V(\theta) = \|\theta - \QS\|_\infty$ is a global Lyapunov function for this algorithm's limiting DI. This immediately implies that this limiting DI has $\{\QS\}$ as its unique globally asymptotically stable equilibrium and, hence, $\theta_n \to \QS$ a.s.
\end{proposition}

\begin{remark}
    Our conclusion that Q-learning converges to $\QS$ is well-known in the literature. What is novel, though, is our DI-based proof. Our analysis reveals that even the dynamics underlying tabular Q-learning with $\epsilon$-greedy exploration is intricate and complex, changing discontinuously from one greedy region to another. The crux of convergence amidst these complexities is a global Lyapunov function rooted in the Bellman optimality operator. This Lyapunov assurance is absent in broader function approximations, explaining their tendency to converge to arbitrary invariant sets. 
\end{remark}

\section{Proofs}
\label{s:Proofs}

Section~\ref{s:Setup.Main.Result} results are proved here: proofs of Theorem~\ref{thm:main.result} and Corollary~\ref{cor:fixed.points.proj.Bellman} are in Section~\ref{ss:proof.main.result}, that of Proposition~\ref{prop:tabular} in Section~\ref{ss:proof.tabular.lyapunov}, and those of Lemmas~\ref{lem:base.driving.function} and \ref{lem:htheta} in Section~\ref{ss:proof.lemmas}.

\subsection{Proof of Main Results (Theorem~\ref{thm:main.result} and Corollary~\ref{cor:fixed.points.proj.Bellman})}
\label{ss:proof.main.result}
Using our Step 1 from Section~\ref{s:appl}, recall that linear Q-learning and linear SARSA(0)  from  \eqref{e:generic.update} can be rewritten as \eqref{e:SA.update.rule}, where $f,$ $\rho_n,$ and $M_{n + 1}$ are given by \eqref{e:f.Defn} (or, equivalently, the expression in Lemma~\ref{lem:base.driving.function}), \eqref{e:rhon.defn}, and \eqref{e:Mn.defn}, respectively. Compared to a standard stochastic approximation  \cite{robbins1951stochastic, borkar2022stochastic, benaim1999dynamics}, the analysis of \eqref{e:generic.update} has two main challenges. First, the resultant driving function $f$ is discontinuous (see Lemma~\ref{lem:base.driving.function}), a consequence of the $\epsilon$-greedy exploration. Second, the perturbation term $\rho_n$ need not necessarily decay to $0,$ especially when $(\theta_n)$ continually jumps between two or more greedy regions. Recall that $\rho_n$ arises due to the experience-replay and target-network-based sampling of $s_n, a_n, s_n',$ and $a_n'.$

Our approach to overcoming the above two challenges is as follows. We handle the discontinuity by treating \eqref{e:generic.update} as a Stochastic Recursive Inclusion (SRI) \cite{benaim2005stochastic, borkar2022stochastic}, a framework that admits discontinuous driving functions---unlike classical stochastic approximation. This enables us to study the limiting dynamics of \eqref{e:generic.update} using the powerful DI viewpoint instead of the standard differential-equation-based one. Separately, we handle $\rho_n$ by carefully decomposing it into terms that arise only due to experience replay, and those that arise due to the target network. On a sample path where the $(\theta_n)$ iterates are stable, we exploit the fact that the target network parameter $\thetam_n$ is updated using a faster timescale to show that $\|\thetam_n - \theta_n\|$ and, hence, the terms that depend on $\thetam_n,$ asymptotically vanish. In contrast, for the terms that arise due to experience replay, we show that a telescopic sum exists that ensures their cumulative effect is asymptotically negligible. 

A key result that we build upon to handle the discontinuity of $f$ in \eqref{e:generic.update} is \cite[Corollary~5.1]{borkar2022stochastic}, which concerns the convergence of Stochastic Recursive Inclusions (SRIs). To help the reader, we first describe SRIs and then state the above result. Alongside, we also explain why this result is not sufficient to directly prove  Theorem~\ref{thm:main.result}. Finally, we prove Theorem~\ref{thm:main.result}.

An SRI is a generic update like
\begin{equation}
\label{e:stochastic.recursive.inclusion}
    \theta_{n + 1} = \theta_n + \alpha_n[y_n +  M_{n + 1}], \qquad n \geq 0,
\end{equation}
where $y_n$ is some desired update vector satisfying $y_n \in h(\theta_n)$ for some set-valued map $h,$ $\alpha_n$ is some stepsize, and $M_{n + 1}$ is noise. A stochastic approximation is a special case of an SRI, where $h(\theta)$ is a singleton for all $\theta.$ Note that \eqref{e:generic.update} has the form given in \eqref{e:stochastic.recursive.inclusion}, but with an additional perturbation term $\rho_n.$ In particular, in the case of \eqref{e:generic.update}, $y_n = f(\theta_n) \in h(\theta_n),$ where $f$ and $h$ are as in \eqref{e:f.Defn} and \eqref{e:filippov.stitch}, respectively.

We next state \cite[Corollary~5.1]{borkar2022stochastic}, which provides a sufficient set of conditions for the $(\theta_n)$ sequence generated by \eqref{e:stochastic.recursive.inclusion} to converge to the invariant sets of the DI $\dot{\theta}(t) \in h(\theta(t)).$ 

\begin{theorem}[Corollary~5.1, \cite{borkar2022stochastic}]
\label{thm:SRI.convergence}

\noindent Consider a generic SRI like  \eqref{e:stochastic.recursive.inclusion} and suppose the following conditions hold.
{
\renewcommand{\theenumi}{$\cC_\arabic{enumi}$}
\begin{enumerate}
    \item \label{a:Set.Lipschitz} \textbf{Driving function}: $h$ is Marchaud or continuous in a set-value sense, i.e., 
    \begin{enumerate}
        \item $h(\theta)$ is convex and compact for all $\theta \in \bR^d;$
    
        \item $\exists K_h > 0$ such that $
        \sup_{y \in h(\theta)} \|y\| \leq K_h(1 + \|\theta\|)$ for all $\theta \in \bR^d,$ and
    
        \item $h$ is upper semicontinuous or, equivalently,  $\{(\theta, y) \in \bR^d \times \bR^d: y \in h(\theta)\}$ is closed.
    \end{enumerate}

    \item \label{a:Robbins} \textbf{Stepsize schedule}: $(\alpha_n)$ is a non-increasing sequence that satisfies the Robbins-Monro condition, i.e., $\sum_{n = 0}^\infty \alpha_n = \infty,$ but $\sum_{n = 0}^\infty \alpha^2_n < \infty.$
    
    \item \label{a:Noise} \textbf{Noise behavior}: $(M_n)$ is a square-integrable martingale-difference sequence adapted to an increasing family of $\sigma$-fields $(\cF_n).$ Also, there exists a constant $K_m \geq 0$ such that, for all $n \geq 0,$ 
    \[
        \bE[\|M_{n + 1}\|^2 |\cF_n] \leq K_m [1 + \|\theta_n\|^2]\ \textnormal{a.s.}\ 
    \]
\end{enumerate}
}
\noindent Then, a.s. on every sample path where the iterate sequence $(\theta_n)$ is  stable, i.e., $\sup_{n \geq 0}\|\theta_n\| < \infty,$ we have that $(\theta_n)$ converges to a (possibly sample-path dependent) closed, connected, internally chain transitive set of $\dot{\theta}(t) \in h(\theta(t)).$
\end{theorem}

Theorem~\ref{thm:SRI.convergence} is not directly applicable to \eqref{e:generic.update} due to the additional perturbation term, $\rho_n$. Instead, we prove Theorem~\ref{thm:main.result} by building upon Theorem~\ref{thm:SRI.convergence}'s proof from \cite{borkar2022stochastic}. Our  strategy involves, firstly, verifying that
$h,$ $(\alpha_n),$ and $(M_{n}),$ as defined in the context of \eqref{e:generic.update}, meet the criteria stipulated in Theorem~\ref{thm:SRI.convergence}. Thereafter, we prove that the cumulative impact of the $\rho_n$'s is asymptotically negligible, a key and intricate part of our analysis. Finally, we show that, under the above conditions, the core arguments and thereby the conclusions of Theorem~\ref{thm:SRI.convergence} are upheld, which then leads to our result. 

\noindent \emph{\textbf{Proof of Theorem~\ref{thm:main.result}}}.
We first show that the conditions \ref{a:Set.Lipschitz}, \ref{a:Robbins}, and \ref{a:Noise} of Theorem~\ref{thm:SRI.convergence} hold for \eqref{e:generic.update}. 
    
Consider \ref{a:Set.Lipschitz}.  Lemma~\ref{lem:htheta} shows that $h(\theta)$ is convex. Separately, since $|\supp(\theta)| \leq |\cA^\cS|$ (a finite number), we also have that $h(\theta)$ is closed and bounded (hence, compact). This establishes (\ref{a:Set.Lipschitz}.a). We also  have 
\[
    \sup_{y \in h(\theta)} \|y\| \leq  \max\left\{ \max_{\ba \in \cA^\cS} \|b_\ba\|, \max_{\ba \in \cA^\cS}  \|A_\ba\|\right\} (1 +  \|\theta\|).  
\]
Separately, since there are finite states and actions, we have that $\exists K_\phi, K_r \geq 0$ such that, for any $s, s' \in \cS$ and $a \in \cA,$
\begin{equation}
    \label{e:reward.phi}
    \|\phi(s, a)\| \leq K_\phi \qquad \text{ and } \qquad |r(s, a, s')| \leq K_r.
\end{equation}
From \eqref{e:b_ba.defn}, \eqref{e:A_ba.defn}, and \eqref{e:reward.phi}, it then follows that $\|b_\ba\| \leq K_\phi K_r$ and $\|A_\ba\| \leq (1 + \gamma \sqrt{|\cS|}) K_\phi^2.$ Hence, (\ref{a:Set.Lipschitz}.b) is satisfied for $K_h := K_\phi \max\{K_r, (1+ \gamma \sqrt{|\cS|}) K_\phi\}.$ It remains to establish the upper semicontinuity of $h.$ That is, for any sequences $(x_n)$ and $(z_n)$ such that $x_n \to \theta,$ $z_n \to y,$ and $z_n \in h(x_n) \, \forall n \geq 0,$ we need to show that $y \in h(\theta).$ This is a consequence of the `Filippov convexification' and, hence, we use the form of $h$ given in \eqref{e:filippov.stitch} for deriving it. Let $\delta > 0$ be arbitrary. Then, $\exists N_\delta \geq 0$ such that $x_n \in B(\theta, \delta)$ for all $n \geq N_\delta.$ Further, for each such $n,$ since $B(\theta, \delta)$ is open, there is also small ball around $x_n$ that is contained in $B(\theta, \delta)$ which, in turn, implies 
\begin{equation}
\label{e:yn.superset}
z_n \in h(x_n) \subseteq \overline{\co}(f(B(\theta, \delta))).
\end{equation}
Because the set on the extreme right is closed and $y$ is the limit of $(z_n),$ we have $y \in \overline{\co}(f(B(\theta, \delta))).$ The choice of $\delta$  being arbitrary finally shows that $y \in h(\theta),$ as desired. 

Condition \ref{a:Robbins} holds due to our stepsize assumption in \ref{a:stepsize}. 

Next consider \ref{a:Noise}. Let $\cF_n$ be the $\sigma$-field generated by $\thetam_0, \theta_0, \ldots, \theta_{-\ell}, s_0, a_0, s_0', a_0', \ldots, s_{n - 1}, a_{n - 1},$ $s_{n - 1}', a_{n - 1}'.$ Then, from \eqref{e:Mn.defn}, it follows that $(M_{n})$ is a martingale-difference sequence adapted to $(\cF_n).$ We next show that $(M_{n})$ is square integrable, i.e., $\bE\|M_{n}\|^2 < \infty$ for all $n \geq 1.$ For $n \geq 0,$  \eqref{e:deltan.defn} and \eqref{e:reward.phi} show that
\begin{align}
    \|\delta_n \phi(s_n, a_n)\| \leq {} &  |\delta_n| \, \|\phi(s_n, a_n) \| \nonumber \\
    \leq {} & K_\phi \left[K_r + \gamma K_\phi \|\thetam_n\| +  K_\phi \|\theta_n\|\right], \label{e:delta.phi.Bd}
\end{align}
which, when combined with \eqref{e:Mn.defn}, gives  
\begin{equation}
\label{e:M_n.Bd}
\|M_{n + 1}\| \leq 2K_\phi \left[K_r + \gamma K_\phi \|\thetam_n\| +  K_\phi \|\theta_n\|\right].
\end{equation}
Separately, from \eqref{e:target.network.update}, we have that
\begin{equation}
\label{e:thetam_n.Bd}
    \|\thetam_{n + 1}\| \leq (1 - \tau_n \zeta_{n + 1}) \|\thetam_n\| + \zeta_{n + 1} \tau_n \|\theta_n\|.
\end{equation}
Since $\|\theta_0\|^2, \|\thetam_0\|^2 < \infty,$ we have from \eqref{e:generic.update} and \eqref{e:target.network.update}, and the $n = 0$ case of  \eqref{e:delta.phi.Bd}, \eqref{e:M_n.Bd}, and \eqref{e:thetam_n.Bd}, that $\bE\|M_1\|^2 < \infty,$ $\bE\|\theta_1\|^2 < \infty,$ and $\bE\|\thetam_1\|^2 < \infty.$  The desired result then follows by induction. It remains to establish the condition on $\bE [\|M_{n + 1}\|^2 |\cF_n].$ However, we gets this for free from \eqref{e:M_n.Bd}; specifically, for $K_m = 12 K_\phi^2 \max\{K_r^2, K_\phi^2\},$ we have that
\begin{equation}
\label{e:E.|M_{n + 1}|^2.Bd}
    \bE[\|M_{n + 1}\|^2 |\cF_n] \leq K_m[1 + \|\thetam_n\|^2 + \|\theta_n\|^2].
\end{equation}

Note that there is an additional $\|\thetam_n\|^2$ term in the last inequality above compared to the one required in \ref{a:Noise}. We now show that this additional term does not pose any issues and can be handled similarly to how the $\|\theta_n\|$ term is dealt with in Theorem~\ref{thm:SRI.convergence}'s proof in \cite{borkar2022stochastic}. Specifically, in Theorem~\ref{thm:SRI.convergence}'s proof, \ref{a:Noise} is used to show that the sequence $(\zeta_n),$ where $\zeta_{n} = \sum_{k = 0}^{n - 1} \alpha_k M_{k + 1},$ converges a.s. on the event $\{\sup_n \|\theta_n\| < \infty\}.$ In particular, the condition on $\bE [\|M_{n + 1}\|^2 |\cF_n]$ from \ref{a:Noise} and the square-integrability of $(\alpha_n)$ due to \ref{a:Robbins} is used to trivially show that $\sup_{n \geq 0} \|\theta_n\| < \infty$ implies $\sum_{n \geq 0} \alpha_n^2 \bE [\|M_{n + 1}\|^2|\cF_n] < \infty.$ Theorem~C.3 from \cite{borkar2022stochastic} is then invoked to show this latter conclusion is sufficient for $(\zeta_n)$'s convergence, as desired. In our setting, from \eqref{e:thetam_n.Bd} and a simple inductive argument, we have that $\sup_{n \geq 0} \|\thetam_n\| < \infty$ a.s. on the event $\{\sup_n \|\theta_n\| < \infty\}.$ Hence, the same arguments as in \cite{borkar2022stochastic}, along with \eqref{e:E.|M_{n + 1}|^2.Bd}, can again be used to show that, a.s. on $\{\sup_{n \geq 0} \|\theta_n\| < \infty\},$ we have that $\sum_{n \geq 0} \alpha_n^2 \bE [\|M_{n + 1}\|^2|\cF_n] < \infty$ and, hence, $(\zeta_n)$ converges. 

We now analyze the asymptotic behavior of $(\theta_n)$ generated using \eqref{e:generic.update}. Fix a sample point where $\sup_{n \geq 0} \|\theta_n\| < \infty.$ Then, as discussed above, we have that $\sup_{n \geq 0} \|\thetam_n\| < \infty.$

First, we look at $\thetam_n$'s asymptotic behavior. Clearly,  \eqref{e:generic.update} and \eqref{e:target.network.update} can be jointly viewed as a two-timescale algorithm. Specifically, since $\alpha_n/\tau_n \to 0,$ it follows that $(\theta_n)$ is updated on a slower timescale relative to $(\thetam_n);$ hence, the $(\theta_n)$ sequence would appear static from the viewpoint of \eqref{e:target.network.update}. Now, for the case where $\theta_n \equiv \theta$ for some $\theta \in \bR^d,$ \eqref{e:target.network.update}'s  limiting ODE is $\dot{x}(t) = \Delta(\theta - x(t)),$ where $\Delta > 0$ is as defined above \eqref{e:target.network.update}. Importantly, this ODE has $\theta$ as its globally asymptotically stable equilibrium.  This limit, as a function of $\theta,$ is trivially Lipschitz continuous. Hence, from \cite[Lemma~8.1]{borkar2022stochastic}, we have
\begin{equation}
\label{e:target.network.tracks.theta.n}
    \|\thetam_n - \theta_n\| \to 0.
\end{equation}

Next, we discuss $(\theta_n)$'s asymptotic behavior. Let $T > 0$ be an arbitrary horizon. Further, for $n \geq 0,$ let $m_n$ be the smallest index $k$ such that $T \leq \sum_{j = n}^{n + k} \alpha_j \leq T + 1.$ Then, for any $n \geq 0$ and $m$ such that $0 \leq m \leq m_n,$ we have that
\[
% \label{e:theta.n+m+1.decomposition}
%
    \theta_{n + m + 1} = \theta_n + \sum_{j = n}^{n + m} \alpha_j f(\theta_n) + \sum_{j = n}^{n + m} \alpha_j M_{j + 1} + \sum_{j = n}^{n + m} \alpha_j \rho_j.
\]
The above expression is of the form given in \cite[(2.1.6)]{borkar2022stochastic}\footnote{Due to the discontinuity in $f,$ we actually need to consider the analogous form needed to derive \cite[Lemma~5.1]{borkar2022stochastic}; however, as stated in ibid, the latter's proof mimics that of \cite[Lemma~2.1]{borkar2022stochastic}.}, except for the additional sum involving the perturbation terms. Hence, if we can show that $\sup_{0 \leq m \leq m_n } \|\sum_{j = n}^{n + m} \alpha_j \rho_j\| \to 0$ as $n \to \infty,$ then Theorem~\ref{thm:main.result} would follow by using similar arguments as in \cite[Corollary~5.4]{borkar2022stochastic}. 

We now show the above claim. For this, we decompose $\alpha_n \rho_n$  into terms that arise due to experience replay and those arise due to the target network. To begin with, we have from \eqref{e:v.defn}, \eqref{e:deltan.defn}, \eqref{e:b_ba.defn}, and \eqref{e:A_ba.defn} that, for any $\thetam, \theta^{(0)}, \ldots, \theta^{(-\ell)},$ 
\begin{align*}
    v (\thetam, \theta^{(0)}, \ldots, \theta^{(-\ell)})
    = {} & \bE\left[\delta_0 \phi(s_0, a_0) \middle|\thetam_0 = \thetam, \theta_{k} = \theta^{(k)}, k = -\ell, \ldots, 0\right] \nonumber \\
    = {} & \sum_{k = 0}^\ell \mu_k  \bigg( \sum_{\ba \in \cA^{\cS}} \Big[(b_\ba - A_\ba \theta^{(0)}) \indc[\theta^{(-k)} \in \cR_\ba] - \gamma \Phi^\tr D_\ba^\epsilon P_{\ba}^{\epsilon'} \Phi \theta^{(0)} \indc[\theta^{(-k)} \in \cR_\ba] \Big]\\
    {} & \hspace{0.5em} + \sum_{\ba, \ba' \in \cA^{\cS}}\gamma \Phi^\tr D_\ba^\epsilon P_{\ba'}^{\epsilon'} \Phi \thetam \indc[\theta^{(-k)} \in \cR_\ba, \thetam \in \cR_{\ba'}] \bigg).
\end{align*}
Therefore, using \eqref{e:rhon.defn}, we get that 
\begin{align}
    \alpha_n \rho_n = {} & \alpha_n \left[v(\thetam_n, \theta_n, \ldots, \theta_{n - \ell}) - f(\theta_n) \right] \nonumber \\
    %
    % = {} & \alpha_n \sum_{k = 0}^\ell  \mu_k \bigg( \sum_{\ba \in \cA^\cS}  (b_\ba - A_\ba \theta_n) \indc[\theta_{n - k} \in \cR_\ba] +\nonumber \\
    %
    % {} & \hspace{0.25em} \sum_{\ba, \ba' \in \cA^{\cS}} \gamma \Phi^\tr D_\ba^\epsilon \Big[P_{\ba'}^{\epsilon'} \Phi \thetam_n - P_{\ba}^{\epsilon'} \Phi \theta_n\Big] \indc[\theta_{n - k} \in \cR_\ba, \thetam_n \in \cR_{\ba'}] \nonumber \\
    %
    % {} & \hspace{0.75em} \times \indc[\theta_{n - k} \in \cR_\ba, \thetam_n \in \cR_{\ba'}] - f(\theta_n)\bigg) \nonumber \\
    %
    % {} & - (b_\ba - A_\ba \theta_n) \indc[\theta_n \in \cR_\ba] \bigg] \nonumber \\
    %
    = {} & \rho_n^{(1)} + \rho_n^{(2)} + \rho_n^{(3)} + \rho_n^{(4)}, \label{e:alpha.rho.n.decomposition}
\end{align}
where 
\begin{align}
    \rho_n^{(1)} := 
    % {}  & \sum_{k = 0}^\ell \mu_k \sum_{\ba \in \cA^\cS} \bigg[ \alpha_{n - k} (b_\ba - A_\ba \theta_{n - k}) \indc[\theta_{n - k}  \in \cR_\ba] \nonumber \\
    %
    % {} & - \alpha_n (b_\ba - A_\ba \theta_n) \indc[\theta_n \in \cR_\ba] \bigg], \nonumber  \\
    %
    % = 
    {} & \sum_{k = 0}^\ell \mu_k \left[\alpha_{n - k} f(\theta_{n - k}) - \alpha_n f(\theta_n) \right]; \label{e:rho_1.telescopic.sum}\\[2ex]
    \rho_n^{(2)} := {} & \sum_{k = 0}^\ell \mu_k  (\alpha_{n} - \alpha_{n - k}) f(\theta_{n - k}); \label{e:rho_2.telescopic.sum}\\[2ex]
    \rho_{n}^{(3)} := {} & \alpha_n \sum_{k = 0}^\ell  \mu_k \bigg(\sum_{\ba \in \cA^\cS} \bigg[(b_\ba - A_\ba \theta_n) \indc[\theta_{n - k} \in \cR_\ba] \nonumber\\
    {} & \hspace{1em} +  \gamma  \Phi^\tr D_\ba^\epsilon P_{\ba}^{\epsilon'} \Phi(\theta_{n - k}  - \theta_n) \indc[\theta_{n - k} \in \cR_\ba] \bigg] - f(\theta_{n - k})\bigg) \nonumber \\
    = {} & \alpha_n \sum_{k = 0}^\ell  \mu_k  \sum_{\ba \in \cA^\cS} \left[A_\ba + \gamma  \Phi^\tr D_\ba^\epsilon P_{\ba}^{\epsilon'} \Phi \right] (\theta_{n - k}  - \theta_n) \indc[\theta_{n - k} \in \cR_\ba]; \label{e:rho_3.telescopic.sum}  \\
    \rho_n^{(4)} := {} & \alpha_n \sum_{k = 0}^\ell \mu_k  \hspace{-0.25em}\sum_{\ba, \ba' \in \cA^{\cS}} \hspace{-0.1em} \gamma \Phi^\tr D_\ba^\epsilon \Big[P_{\ba'}^{\epsilon'} \Phi \thetam_n - P_{\ba}^{\epsilon'} \Phi \theta_{n- k}\Big] \indc[\theta_{n - k} \in \cR_\ba, \thetam_n \in \cR_{\ba'}]. \label{e:rho_4.telescopic.sum} 
\end{align}
%
%
% \begin{figure*}[ht!]
% \normalsize
% \begin{align}
%     \rho_n^{(1)} &:= \sum_{k = 0}^\ell \mu_k \left[\alpha_{n - k} f(\theta_{n - k}) - \alpha_n f(\theta_n) \right]; \label{e:rho_1.telescopic.sum}\\[2ex]
%     %
%     \rho_n^{(2)} &:= \sum_{k = 0}^\ell \mu_k  (\alpha_{n} - \alpha_{n - k}) f(\theta_{n - k}); \label{e:rho_2.telescopic.sum}\\[2ex]
%     %
%     \rho_{n}^{(3)} &:= \alpha_n \sum_{k = 0}^\ell  \mu_k \bigg(\sum_{\ba \in \cA^\cS} \Big[ b_\ba - A_\ba \theta_n + \gamma  \Phi^\tr D_\ba^\epsilon P_{\ba}^{\epsilon'} \Phi(\theta_{n - k}  - \theta_n)\Big] \indc[\theta_{n - k} \in \cR_\ba]  - f(\theta_{n - k}) \bigg) \nonumber \\[2ex]
%     %
%     & =  \alpha_n \sum_{k = 0}^\ell \mu_k  \sum_{\ba \in \cA^\cS} \left[A_\ba + \gamma  \Phi^\tr D_\ba^\epsilon P_{\ba}^{\epsilon'} \Phi \right](\theta_{n - k}  - \theta_n) \indc[\theta_{n - k} \in \cR_\ba]; \label{e:rho_3.telescopic.sum}\\[2ex]
%     %
%     \rho_n^{(4)} & := \alpha_n \sum_{k = 0}^\ell \mu_k  \sum_{\ba, \ba' \in \cA^{\cS}} \gamma \Phi^\tr D_\ba^\epsilon \Big[P_{\ba'}^{\epsilon'} \Phi \thetam_n - P_{\ba}^{\epsilon'} \Phi \theta_{n- k}\Big] \indc[\theta_{n - k} \in \cR_\ba, \thetam_n \in \cR_{\ba'}]. \label{e:rho_4.telescopic.sum}
% \end{align}
% \end{figure*}
%
The sum in \eqref{e:alpha.rho.n.decomposition} is our proposed decomposition for $\alpha_n \rho_n$. Note that $\rho_n^{(1)}, \rho_n^{(2)},$ and $\rho_n^{(3)}$ appear due to experience replay, while $\rho_n^{(4)}$'s origin is mainly due to the target network.

We now individually study the behaviors of the four terms in the above decomposition. We first look at the $\rho_n^{(4)}$ term. For any $(s, a)$  and any $\theta \in \cR_\ba$, $P^{\epsilon'}_\ba$'s definition shows that
\begin{multline}
\label{e:P.epsp.defn}
    P_{\ba}^{\epsilon'}(\cdot|s, a) \Phi \theta  =   \sum_{s' \in \cS} \bP(s'|s, a)\bigg[(1 - \epsilon')\max_{b} \phi^\tr (s', b) \theta + \frac{\epsilon'}{|\cA|} \sum_{a' \in \cA}  \phi^\tr (s', a') \theta\bigg].
\end{multline}
Separately, for any $\theta, \theta',$
\begin{equation}
\label{e:max.gap}
    |\max_b \phi^\tr (s',b) \theta -  \max_b \phi^\tr (s',b) \theta'| 
    \leq \max_{b}|\phi^\tr(s',b) \theta - \phi^\tr(s',b) \theta'|. 
\end{equation} 
Therefore, 
\begin{align*}
    \Big\| P_{\ba'}^{\epsilon'} \Phi \thetam_{n} - P_{\ba}^{\epsilon'} \Phi \theta_{n - k}  \Big\|_\infty \indc[\thetam_n \in \cR_{\ba'}, \theta_{n - k} \in \cR_\ba] 
    \leq {} &  \|\Phi \theta_{n - k} - \Phi \thetam_{n}\|_\infty \\
    \leq {} & \|\Phi (\theta_n - \thetam_n)\|_\infty + \|\Phi(\theta_n - \theta_{n - k})\|_\infty,
\end{align*}
where the first relation follows from \eqref{e:P.epsp.defn} and \eqref{e:max.gap}, while the second is due to triangle inequality. Now, for any $0 \leq k \leq \ell,$
\begin{align*}
    \|\theta_n - \theta_{n - k}\| \leq {} & \sum_{j = n - \ell}^{n - 1}\| \theta_{j + 1} - \theta_{j}\| \\
    \overset{(a)}{\leq} {} & \sum_{j = n - \ell}^{n - 1} \alpha_j |\delta_j| \ \|\phi(s_j, a_j)\| \\
    \overset{(b)}{=} {} & O(\alpha_{n - \ell}),
\end{align*}
where (a) is due to \eqref{e:generic.update}, while (b) follows since $(\alpha_n)$ is monotonically decreasing (see \ref{a:stepsize}),  the rewards and feature vectors are bounded (see \eqref{e:reward.phi}), and both $\sup_{n \geq 0} \|\theta_n\|$ and $\sup_{n \geq 0} \|\thetam_n\|$ are finite. Therefore, 
\[
    \frac{\|\rho_n^{(4)}\|}{\alpha_n} = O(\|\thetam_n - \theta_n\|) + O(\alpha_{n - \ell}),
\]
which converges to $0$ due to \eqref{e:target.network.tracks.theta.n} and the fact that $\alpha_{n - \ell} \to 0.$ Now, since $\sum_{j = n}^{j = n + m_n} \alpha_j \leq  T + 1,$ we have, as $n \to \infty,$
\[
    \sup_{0 \leq m \leq m_n} \|\sum_{j = n}^{n + m} \rho_j^{(4)}\| = O(T \sup_{j \geq n} \|\thetam_j - \theta_j\|  + T\alpha_{n - \ell}) \to 0.
\]

Next, observe that  $\|\rho_n^{(3)}\|/\alpha_n = O( \|\theta_{n - k} - \theta_n\|).$
Hence, by arguing as above, we have, as $n \to \infty,$
\[
    \sup_{0 \leq m \leq m_n} \|\sum_{k = n}^{n + m} \rho_j^{(3)}\| = O(T \alpha_{n - \ell}) \to 0.
\]

Finally, we study the asymptotic behaviors of $\rho_n^{(1)}$ and $\rho_n^{(2)}.$ Unlike $\rho_n^{(3)}$ and $\rho_n^{(4)},$ though, these terms are not $o(\alpha_n),$ i.e., $\|\rho_n^{(1)}\|/\alpha_n$ and $\|\rho_n^{(2)}\|/\alpha_n$ do not decay to $0.$ However, as we now show, there exists a telescopic sum that ensures their cumulative effect over any finite $T$-length horizon is negligible. Formally, 
\begin{align*}
    \sum_{j = n}^{n + m} \alpha_j \rho_j^{(1)} = {} & \sum_{k = 0}^\ell \mu_k \sum_{j = n}^{n + m}[\alpha_{j - k} f(\theta_{j - k}) - \alpha_j f(\theta_j)] \\
   = {} & \sum_{k = 0}^\ell \mu_k \bigg[\sum_{j = n}^{n + k - 1} \alpha_{j - k} f(\theta_{j - k}) - \sum_{j = n + m + 1 - k}^{n + m} \alpha_j f(\theta_j)\bigg],
\end{align*}
where the intermediate terms get canceled due to their telescopic nature. Note that the two inner summations contain at most $\ell$ many terms. Combining these statements with the facts that $\sup_{n \geq 0} \|\theta_n\| < \infty$ and that $(\alpha_n)$ is non-increasing (see \ref{a:stepsize}), we get $\sup_{m \geq 0} \bigg\|\sum_{j = n}^{n + m} \alpha_j \rho_j^{(1)}\bigg\| = O(\alpha_{n - \ell}).$ Similarly, we have $\sup_{m \geq 0} \|\sum_{j = n}^{n + m} \alpha_j \rho_j^{(2)}\| = O(\alpha_{n - \ell}).$

Hence, $\sum_{0 \leq m \leq m_n} \|\sum_{j = n}^{n + m} \alpha_j \rho_j\| \to 0,$ as desired. 
% By arguing as in the proof of \cite[Corollary~5.4]{borkar2022stochastic}, the desired claim now follows. 
\hfill $\blacksquare$

\noindent \emph{\textbf{Proof of Corollary~\ref{cor:fixed.points.proj.Bellman}}}.
    Consider the first claim. Theorem~\ref{thm:main.result} shows that $\{\thS\}$ must a.s. be an invariant set of its limiting DI. From the definition of an invariant set (see the paragraph above Theorem~\ref{thm:main.result}), it now follows that $0 \in h(\thS),$ as desired. 
    
    For the second claim, we consider two exhaustive cases.  

    \noindent \textbf{Case 1}: $\thS \in \intr(\cR_\ba)$ for some deterministic policy $\ba \in \cA^{\cS},$ where $\intr$ means interior and $\cR_\ba,$ recall, is the greedy region associated with the deterministic policy $\ba.$ 
    
    Since $\thS \in \intr(\cR_\ba),$ Lemma~\ref{lem:htheta} and our first claim above show that $b_\ba - A_{\ba} \thS = 0.$ Hence, 
    \[
        \Phi^\tr D_\ba^{\epsilon} \br = \Phi^\tr D_\ba^{\epsilon} (I - \gamma P^{\epsilon'}_\ba) \Phi \thS
    \]
    or $\Phi^\tr D_\ba^\epsilon  [\br + \gamma P_\ba^{\epsilon'} \Phi \thS] = \Phi^\tr D_\ba^\epsilon \Phi \thS.$ For Q-learning, we have $\epsilon' = 0;$ therefore, the $(s, a)$-th entry of the vector $P_\ba^{\epsilon'} \Phi \thS$ is $ \sum_{s'} P(s'|s, a) \phi^\tr(s', \bar{a}(s')) \thS = \sum_{s'} P(s'|s, a) \max_{a'} \phi^\tr(s', a') \thS,$
    where the equality comes from the fact that $\thS \in \cR_\ba.$ Consequently, $\Phi^\tr D_\ba^\epsilon T \Phi \thS = \Phi^\tr D_\ba^\epsilon \Phi \thS.$ Our Assumption~\ref{a:ergodicity} implies that $\Phi^\tr D_\ba^\epsilon \Phi$ is invertible. Hence, it follows that $        \label{e:proj.relation.intr}
        \Phi (\Phi^\tr D_\ba^\epsilon \Phi)^{-1} \Phi^\tr D_\ba^\epsilon T \Phi \thS =  \Phi \thS.$
        %
    % \end{equation}
    %
    Now if we let $\Pi^\epsilon_\ba = \Phi (\Phi^\tr D_{\ba}^{\epsilon} \Phi)^{-1} \Phi D_{\ba}^\epsilon,$ then $\Pi^\epsilon_\ba Q$ defines the projection of $Q$ onto the column space of $\Phi$ with respect to $\|\cdot\|_{D_\ba^\epsilon}.$ Hence, the above $\thS$ relation  can be equivalently stated as $\Pi^\epsilon_\ba T \Phi \thS = \Phi \thS,$ which proves the desired claim. 
    
\vspace{1ex}

    \textbf{Case 2}: $\thS \in \cap_{i = 1}^m \cR_{\ba_i}$ for an $m \geq 1$ and some deterministic polices $\cP \equiv \{\ba_1, \ldots, \ba_m\} \subseteq  \cA^{\cS}\},$ i.e., $\thS$ is on the boundary of the $m$ greedy regions $\cR_{\ba_1}, \ldots, \cR_{\ba_m}$. 
    
    Since $0 \in h(\thS)$ from our first claim, there exist non-negative weights $\lambda_1, \ldots, \lambda_m$ such that $\lambda_1 + \cdots+ \lambda_m = 1$ and $\sum_{i = 1}\lambda_i (b_{\ba_i} - A_{\ba_i} \thS) = 0.$  Now let $\Pi_\cP^\epsilon = \Phi (\Phi^\tr D_\cP^\epsilon \Phi)^{-1} \Phi^\tr D_\cP^\epsilon,$ where $D_\cP^\epsilon = \sum_{i = 1}^m \lambda_i D_{\ba_i}^\epsilon.$ The inverse here is well defined since  $\Phi^\tr D_{\ba_i}^\epsilon \Phi$ is positive definite for all $i$ (due to Assumption~\ref{a:ergodicity}). Clearly, $\Pi_\cP^\epsilon$ defines projection onto the column space of $\Phi$ with respect to $\|\cdot\|_{D_\cP^\epsilon}.$ By arguing as in Case 1 above, we get that $\Pi_\cP^\epsilon T \Phi \thS = \Phi \thS.$ \hfill $\blacksquare$

\subsection{Proof of V being Tabular Q-learning's Lyapunov Function (Proposition~\ref{prop:tabular})}
\label{ss:proof.tabular.lyapunov}

\noindent \emph{\textbf{Proof of Proposition~\ref{prop:tabular}}}.
For tabular Q-learning, since $\Phi = \bI,$ it follows from \eqref{e:b_ba.defn} and \eqref{e:A_ba.defn} that
\begin{align}
    b_\ba - A_\ba \theta = {} & \Phi^\tr \De_{\ba} \br - \Phi^\tr \De_{\ba} (\bI - \gamma \Pep_\ba) \Phi \theta \nonumber\\
    = {} & \De_{\ba} \left( \br + \gamma \Pep_\ba \theta - \theta \right) = \De_{\ba} \left( T(\theta) - \theta \right), \label{e:tab.Q.learning.b.A}
\end{align}
where $T: \bR^{|\cS| |\cA|} \to \bR^{|\cS| |\cA|}$ is the Bellman optimality operator, i.e., for any $s \in \cS$ and $a \in \cA,$ 
\[
    T(\theta)_{s,a} = \br(s, a) + \gamma \sum_{s' \in \cS} \bP(s'|s, a) \cdot \max_{a' \in \cA} \theta(s', a').
\]
Combining \eqref{e:tab.Q.learning.b.A} with Lemma~\ref{lem:htheta}, it then follows that, for any $\theta$ and any $v \in h(\theta),$ where $h$ defines the limiting DI for tabular Q-learning, we have that $v = \sum_{\ba \in \supp(\theta)} \De_{\ba} \lambda_{\ba}^v \left( T(\theta) - \theta \right) $ with $\lambda_{\ba}^v \geq 0$ and $\sum_{\ba \in \supp(\theta)} \lambda_{\ba}^v = 1.$ If we let $M^v = \sum_{\ba \in \supp(\theta)} \lambda_{\ba}^v \De_{\ba}$, and define the operator $T^v: \bR^{|\cS| |\cA|} \to \bR^{|\cS| |\cA|}$ by $T^v(\theta) = \theta + M^v \left( T(\theta) - \theta \right),$ then we can write $v = T^v(\theta) - \theta$. Note that $T^v(\theta^*) = \theta^*$ because of the Bellman optimality equation $T(\theta^*) = \theta^*$. 

The ergodicity assumption (\ref{a:ergodicity}) and the $\epsilon$-greedy sampling rule with $\epsilon > 0$ ensure that each diagonal matrix $\De_{\ba}$ (and hence the convex combination $M^v$) has minimum eigenvalue at least a positive quantity $p_{\min}> 0$. Since $T$ is $\gamma$-contractive in $\|\cdot\|_\infty$ \cite{bertsekas1996neuro}, it follows that $T^v$ is $\tilde{\gamma}$-contractive in the $\|\cdot\|_\infty$ norm, where $\tilde{\gamma} = 1 - p_{\min} (1-\gamma) \in (0,1)$ \cite[Sec. 6.4]{borkar2022stochastic}. 

Following a calculation similar to\footnote{Specifically, use $T^v$ in place of the function $F$ and $\tilde{\gamma}$ in place of $\alpha$ in the derivation of \cite[Thm. 12.1]{borkar2022stochastic}.} that in the proof of  \cite[Thm. 12.1]{borkar2022stochastic}, and using the contraction and fixed point properties of the map $T^v$, now yields that $V(\theta(t))$ is a strictly decreasing function of $t$ along any non-constant solution trajectory of the DI $\dot{\theta}(t) \in h(\theta(t))$. In particular, we have the stronger result $V(\theta(t)) \leq e^{-(1-\tilde{\gamma})t} \,  V(\theta(0)) =  e^{-p_{\min}(1-\gamma)t} \,  V(\theta(0))$ for all times $t \geq 0$. This shows that $V(\theta) = \|\theta - \QS\|_\infty$ is a Lyapunov function, as desired. \hfill $\blacksquare$

\subsection{Proofs of Technical Results (Lemmas~\ref{lem:base.driving.function} and \ref{lem:htheta})}
\label{ss:proof.lemmas}

\noindent \emph{\textbf{Proof of Lemma~\ref{lem:base.driving.function}}}.
Let $\theta \in \bR^d$ be arbitrary. Since $\{\cR_\ba\}$ partitions $\bR^d,$ we have $\theta \in \cR_\ba$ for some unique policy $\ba.$ Now suppose the initial estimates satisfy $\thetam_0 = \theta_0 = \cdots =  \theta_{-\ell} = \theta.$ Then,  \eqref{e:deltan.defn} shows that
\[
    \delta_0 \phi(s_0, a_0) = \phi(s_0, a_0) r(s_0, a_0, s_0') - \left[\phi(s_0,a_0) \phi^\tr(s_0, a_0) - \gamma \phi(s_0, a_0) \phi^\tr (s_0', a_0')\right] \theta.
\]
Further, $d_k^\epsilon = d_\ba^\epsilon$ for all $k = -\ell, \ldots, 0,$  where $d_k^\epsilon$ and $d_\ba^\epsilon$ are the stationary distributions  defined below \ref{a:ergodicity} and above \eqref{e:b_ba.defn}, respectively. Finally, by recalling how $s_0, a_0, s_0'$ and $a_0'$ are sampled from the discussion below \ref{a:ergodicity}, it follows from  \eqref{e:v.defn}, \eqref{e:f.Defn}, \eqref{e:b_ba.defn}, and \eqref{e:A_ba.defn} that
\begin{align}
    f(\theta) = {} & \bE\left[\delta_0 \phi(s_0, a_0) \middle| \thetam_0 = \theta, \theta_k = \theta, k = -\ell, \ldots, 0\right] \nonumber \\
    = {} & b_\ba - A_\ba \theta. \nonumber
\end{align}
Clearly, the above relation is not influenced by the choice of the buffer-sampling-distribution $\mu.$ The result now follows. \hfill $\blacksquare$

\vspace{1ex}

\noindent \emph{\textbf{Proof of Lemma~\ref{lem:htheta}}}.
Fix $\theta \in \bR^d.$ By definition, $h(\theta)$ is a closed convex set. Our first claim is that $b_\ba - A_\ba \theta \in h(\theta)$ for each $\ba \in \supp(\theta).$ From  $h(\theta)$'s convexity, it will then follow that
\begin{equation}
\label{e:co.h.Subset}
    \hp(\theta) := \co\left\{ b_\ba - A_\ba \theta: \ba \in \supp(\theta) \right\} \subseteq h(\theta).    
\end{equation}
To see the claim, consider an arbitrary $\ba \in \supp(\theta).$ For each $\delta > 0,$ $B(\theta, \delta) \cap \cR_\ba \neq \emptyset$ by the definition of $\supp(\theta);$ take $\theta_\delta \in B(\theta, \delta) \cap \cR_\ba.$ Since $B(\theta, \delta_1) \cap \cR_\ba \subseteq B(\theta, \delta_2) \cap \cR_\ba$ for any $\delta_1 \leq \delta_2,$ it follows that $\{\theta_\delta: 0 < \delta < \delta'\} \subseteq B(\theta, \delta')$ for any $\delta' > 0.$ Hence, $\{b_\ba - A_\ba \theta_\delta: 0 < \delta < \delta'\} \subseteq f(B(\theta, \delta'))$ for any $\delta' > 0.$ Now, since $\lim_{\delta \to 0} \theta_\delta = \theta,$ it follows that $b_\ba - A_\ba \theta \in \overline{f(B(\theta, \delta'))} \subseteq \overline{\co}(f(B(\theta, \delta')))$ for any $\delta' > 0.$ Hence, $b_\ba - A_\ba \theta \in h(\theta),$ as desired.

We now prove that $h(\theta) \subseteq \hp(\theta).$ Suppose not. Then there exists $x \in h(\theta)$ such that $x \notin \hp(\theta).$ Since the latter is a closed set, there in fact exists some $\delta' > 0$ such that $\|x - y\|_2 \geq \delta'$ for all $y \in \hp(\theta).$ Now let $\delta_0 > 0$ be the largest $\delta > 0$ such that $B(\theta, \delta) \cap \cR_\ba \neq \emptyset$ if and only if $\ba \in \supp(\theta)$ (the existence of such a $\delta_0$ for $\theta \neq 0$ can be seen from the definition of $\supp(\theta) $ and the fact that the number of $\ba$'s is finite; for $\theta = 0,$ take $\delta_0 = 
 \infty$).  Pick $\delta$ such that $0 < \delta < \min\{\delta_0, \delta'/(2\max_{\ba} \|A_\ba\|_2)\}.$ Because $x \in h(\theta),$ we have $x \in \overline{\co}(f(B(\theta, \delta))).$ The closure implies there exists $x' := \sum_{i = 1}^m \nu_i (b_{\ba_i} - A_{\ba_i} \theta_i) \in  \co(f(B(\theta, \delta)))$ such that $\|x - x'\| < \delta'/2,$ where $m \geq 1$ and, for $1 \leq i \leq m,$  $\theta_i \in B(\theta, \delta),$ $\ba_i \in \supp(\theta),$ and $\nu_i \in [0, 1]$ with $\sum_{i = 1}^m \nu_i = 1.$ Also, since $\delta < \delta'/(2 \max_\ba \|A_\ba\|_2),$ we have $\|x' - x''\|_2 < \delta'/2$ for $x'' := \sum_{i = 1}^m \nu_i (b_{\ba_i} - A_{\ba_i} \theta) \in \hp(\theta).$ However, this implies $\|x - x''\| < \delta',$ which leads to a contradiction. Hence, it holds that $h(\theta) \subseteq \hp(\theta).$ The desired claim follows. \hfill $\blacksquare$

\begin{figure*}[t!]
\centering
\begin{subfigure}{0.49\textwidth}
    \centering
    \includegraphics[width=66mm
    % 0.75\linewidth
    ]{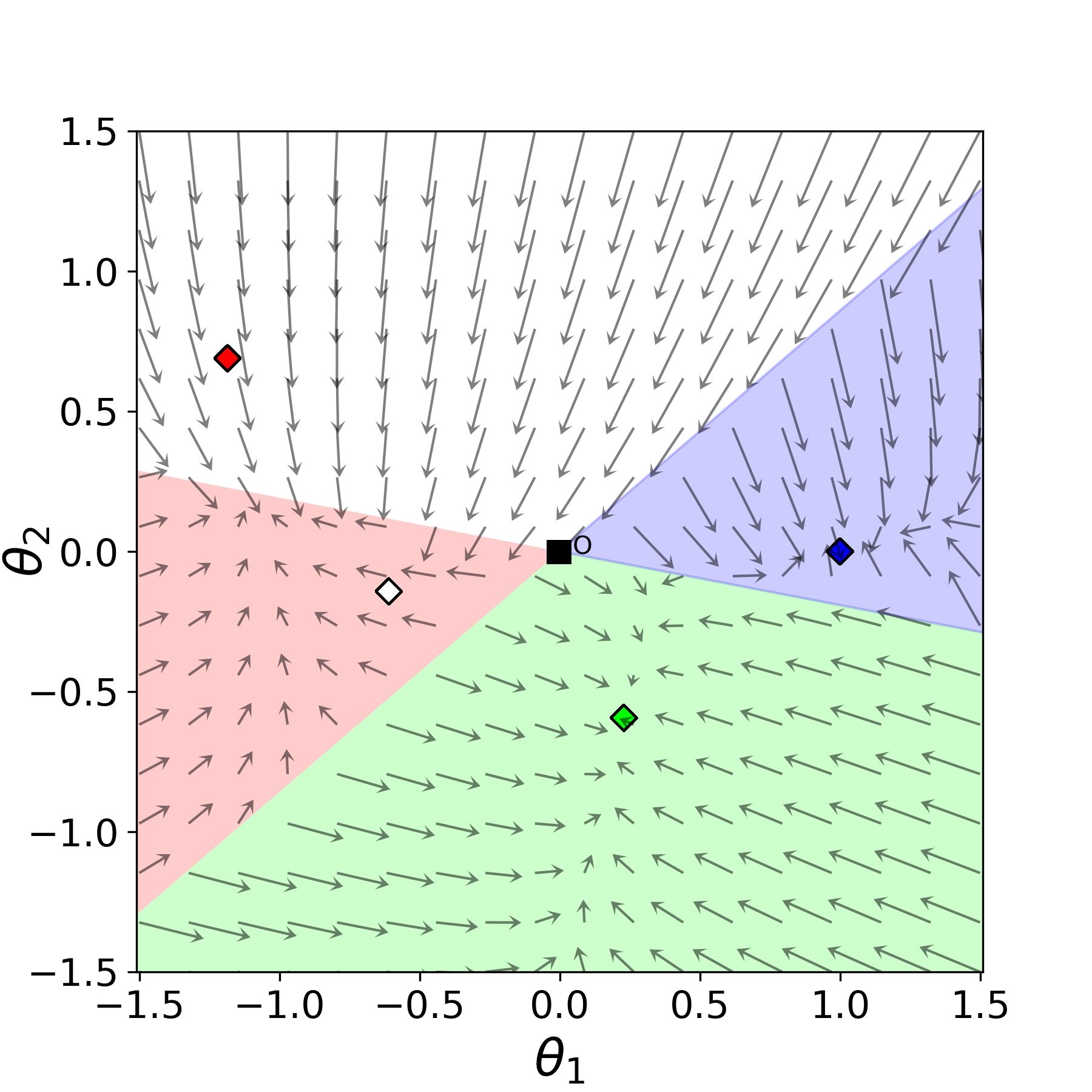}
    \caption{\label{fig:VF.MDP_Intro}}
\end{subfigure}
\begin{subfigure}{0.49\textwidth}
    \centering
    \includegraphics[width=66mm%1.05\linewidth
    ]{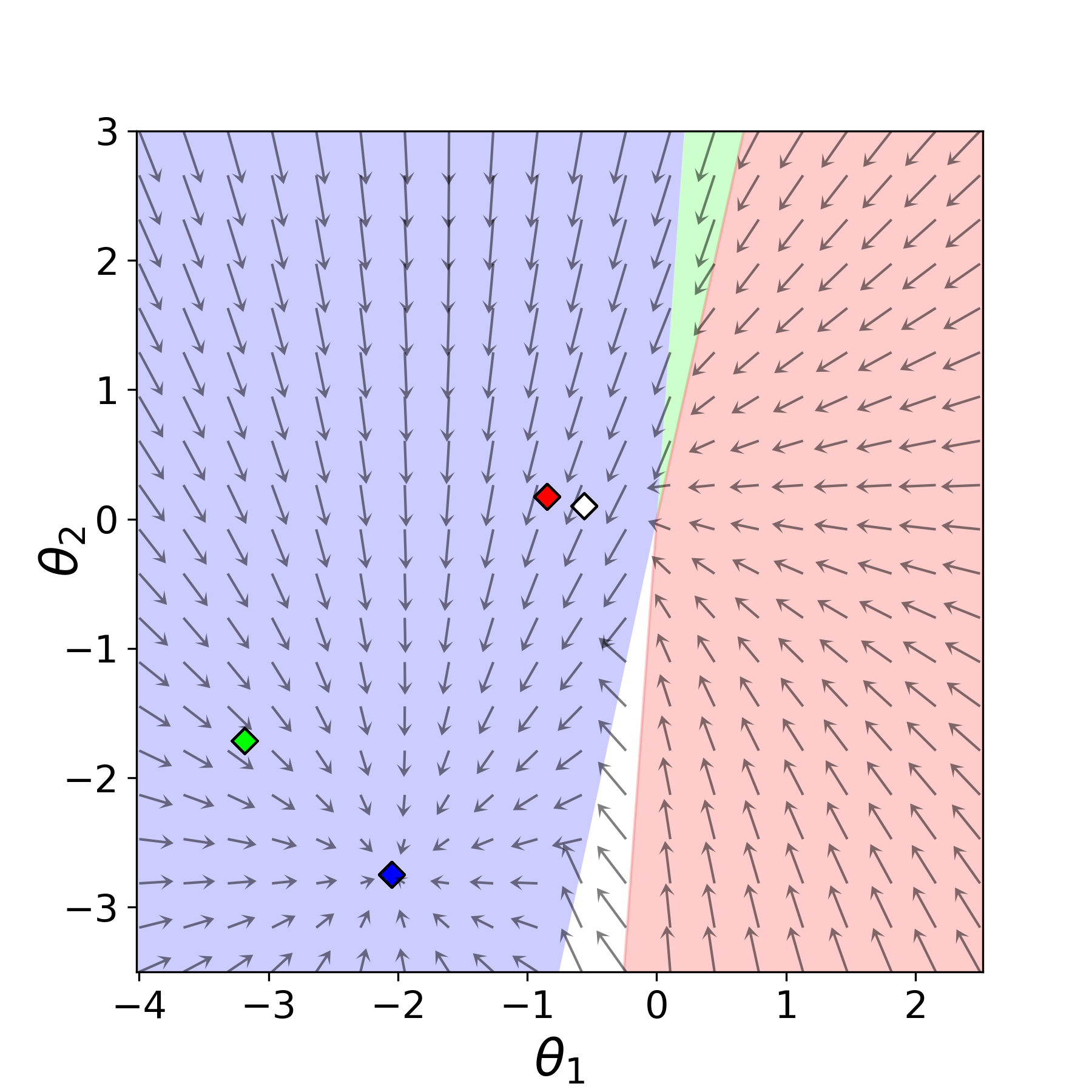}
    \caption{\label{fig:2PartitionOneSelfConsistent}}
\end{subfigure}
\caption{The vector field $f$ for two different MDP settings: (Left) the MDP setting of Figure~\ref{fig:Linear.DQN}, and (Right) an MDP setting where Q-learning will always find the worst policy. The colored regions represent greedy partitions, and the diamond markers are their respective landmarks. The MDP settings can be found in the appendix.}
\label{fig:vector_field}
\vspace{-2ex}
\end{figure*}

\section{Numerical Illustrations}
\label{s:Numerical.Examples}
We now use Theorem~\ref{thm:main.result} and Corollary~\ref{cor:fixed.points.proj.Bellman} to explain the `problematic' behaviors of linear DQN that we saw in Figure~\ref{fig:Linear.DQN}. After that, we discuss an  MDP example where linear DQN always converges to the worst possible policy.

\subsection{Explanation of Linear DQN's Behaviors in Figure~\ref{fig:Linear.DQN}}
We first use Step 1 (Lemma \ref{lem:base.driving.function}) of our framework and the MDP details from the appendix to get the associated vector-field $f: \bR^2 \to \bR^2$. This is given in Figure~\ref{fig:VF.MDP_Intro}. The MDP has four distinct deterministic policies: the colored cones represent the underlying greedy partition $\{\cR_\ba: \ba \in \cA^{\cS}\}$. The local dynamics in each cone is governed by the associated $b_\ba$ and $A_\ba$ values (see \eqref{e:b_ba.defn}, \eqref{e:A_ba.defn}). Each diamond is the point  $A_\ba^{-1} b_\ba,$ the equilibrium for $\dot{\theta}(t) = b_\ba - A_\ba \theta(t)$, which we dub as the \textit{`landmark'} for the dynamics in $\cR_\ba.$ This landmark can be outside $\cR_\ba.$ Each $A_\ba$ is positive definite; hence, the vector field within each cone is oriented towards its landmark. Finally note how the vector field is {\em discontinuous} at the boundaries.

Step 2 convexifies the vector field on the boundaries between regions. In effect, it permits solutions in which the velocity at a boundary point can be {\em any convex combination} of the two velocities associated with the regions comprising the boundary. Applying this logic to Figure \ref{fig:VF.MDP_Intro}, we get the following patterns: {i)} solution trajectories starting from the blue region either remain there and converge to its (blue) landmark
% \footnote{This point corresponds to the star in Figure~\ref{fig:Linear.DQN}.} 
at $[1, 0]^T$ (representing $Q^*$) or cross over to the green region, { ii)} trajectories starting from the green region converge either to its (green) landmark or cross over to the blue region, {iii)} trajectories starting from the red region either cross over to the green region, or hit the red-white boundary in finite time---since the red and white vector fields near the boundary are always oriented towards it, the resultant solutions are forced to `slide' along the boundary towards a point where the red and white regions' velocities oppose each other (a {\em sliding mode attractor}), {iv)} trajectories starting from the white region either cross over to the blue region, after which i) applies, or hit the red-white boundary and slide as before.

Step 3 or Theorem \ref{thm:main.result} guarantees that the idealized variant of Fig.~\ref{fig:Linear.DQN}'s linear DQN converges to a closed, connected, invariant and internally chain transitive set of the above DI. It can rigorously be established\footnote{The  formal proof can be built using references on DIs \cite{aubin2012differential} and discontinuous dynamical systems \cite{cortes2008discontinuous} for rigorous arguments.} 
that the only such sets of this DI are 4 singletons: i) the green and blue landmark points corresponding to a suboptimal and the optimal policy, respectively, ii) a point $\theta_{\text{sliding}}$ on the red-white boundary which is not a proper landmark, and iii) a similar point $\theta_{\text{unstable}}$ on the red-green boundary. This final point, though, is an unstable equilibrium point, because any neighborhood around it contains points from where the DI's solutions will escape away from it. A direct check shows that $0 \in h(\theta)$ holds only at the four points listed above. In fact, repeating the argument from Corollary~\ref{cor:fixed.points.proj.Bellman} confirms that these four points are the only solutions to the projected Bellman fixed-point equation. The projection norm, however, varies from point to point. 

% Thus, this condition is analogous to the notion of $0$ belonging to the subdifferential of a function in the optimization context, wherein it characterizes a critical point.

Figure \ref{fig:DQN.Trajectories} can now be explained as follows: the blue (resp. green) trajectory in Fig.~\ref{fig:DQN.Trajectories} converges to the blue (resp. green)  diamond in Fig.~\ref{fig:VF.MDP_Intro}; recall that the blue diamond is $\QS$'s parameter. In contrast, the red trajectory goes to the sliding-mode attractor on the boundary between the red and white regions. In the last case, the iterates continuously `chatter' or bounce between the red and white regions, which explains the policy oscillation we see in Fig.~\ref{fig:Chattering}. 

% Suppose we define the neighborhood of a stable equilibrium as the set of cones which i) contain the equilibrium, or ii) shares a boundary with the cone(s) containing the equilibrium. That is, the set of places that linear DQN can potentially explore before reaching this equilibrium, e.g., green landmark's neighborhood consists of the green, blue, and red cones. Then, it follows that neither the green nor the sliding-mode attractor is locally optimal! 

% We conjecture that the above understanding can be extended to interpret the DQN behavior seen in Figure~\ref{fig:Scatterplot.DQN}.

\subsection{Reliable Convergence but to the Worst Policy}
Fig~\ref{fig:2PartitionOneSelfConsistent} provides the vector field for linear DQN's limiting DI in the context of another $2$-state, $2$-action MDP (see the appendix for details). Interpreting this vector field as above, we get that linear DQN's iterates a.s. converges to the blue diamond. The striking fact is that the greedy policy associated to this landmark is the {\em worst} of all the 4 deterministic policies, demonstrating a hopeless `no-improvement' scenario. A similar observation has been made in \cite{young2020understanding} for an episodic (finite-horizon, undiscounted) MDP setting. 

\section{Discussion and Future Directions}

On a somber note, our insights about Q-learning and SARSA under arguably the simplest possible (linear) function approximation with $\epsilon$-greedy exploration cast doubt on their utility in more complicated, nonlinear approximation architectures. Unless the specific setting where the algorithms are applied has favorable structural properties (in terms of its limiting DI), the practitioner must anticipate unreliable behaviors. Our work also reinforces the fact that merely ensuring stability of an incremental RL algorithm's iterates is by no means sufficient to guarantee good performance---the discontinuous policy update and the sampling distribution can still induce complex behaviors. 

On the positive side, our approach provides a systematic design pathway for reliable RL algorithms: Ensure the associated DI's attractors lie in regions associated with high-value policies, potentially via Lyapunov techniques.

\appendix

\section{Appendix: Implementation details for Figures \ref{fig:Scatterplot.DQN}, \ref{fig:Linear.DQN} and \ref{fig:2PartitionOneSelfConsistent}}
\label{s:appendix.MDP.details}

Experimental settings in Figures~\ref{fig:Scatterplot.DQN}, \ref{fig:Linear.DQN}, and \ref{fig:2PartitionOneSelfConsistent} are as follows.
\begin{enumerate}
    \item \textbf{Figure}~\ref{fig:Scatterplot.DQN}
    \begin{itemize}
        \item MDP Details: A 100-MDP population is randomly generated. Each MDP's attributes are as follows:
        \begin{enumerate}
            \item $|\cS| = |\cA| = 10,$ i.e., 10 states and 10 actions
            \item Transition probabilities $\bP(s'|s, a)$: Sampled in an IID fashion from  Uniform$[0,1]$, followed by normalization to ensure that $\sum_{s'} \bP(s'|s, a) = 1.$
            \item Reward $r(s,a)$: \hspace{-0.25em} IID sampled from Uniform$[0,1].$
            \item Discount factor $\gamma = 0.75.$
        \end{enumerate}
        
        \item DQN implementation details: 
        \begin{enumerate}
            \item Stepsize: $\alpha_n = \frac{\log n}{10n}.$
            \item Exploration parameter $\epsilon = 0.2$            
            \item Trajectory length: 5000.
            \item Q-value network: Input layer (width $S = 10$) followed by a hidden layer (width $1$) followed by an output layer (width $A = 10$), with ReLU nonlinearity.
            \item Replay buffer size: 100; Batch size: 4.
            \item Target network update period: 4.
        \end{enumerate}
        
    \end{itemize}
    \item \textbf{Figure}~\ref{fig:Linear.DQN}
    \begin{itemize}
        \item MDP Details
        \begin{enumerate}
            \item $|\cS| = |\cA| = 2,$ i.e., 2 states and 2 actions.

            \item Transition matrix \\
            $\{\bP(s'|s, a_1)\}_{s, s'} = \left[ 
                \begin{array}{cc}
                    0.380 & 0.620 \\
                    0.786 & 0.214
                \end{array} \right]$, \\[1ex]
            $\{\bP(s'|s, a_2)\}_{s, s'} = \left[ \begin{array}{cc}
            0.124 & 0.876 \\
            0.426 & 0.574
            \end{array} \right].$

            \item Reward vector \\
            $r = \left[ \begin{array}{c}
                    -0.031 \\
                    0.785 \\
                    -0.282 \\
                    -0.418
                    \end{array} \right].$

            \item Discount factor $\gamma = 0.9.$
        \end{enumerate}

        \item Feature matrix \\
        
        $\Phi = \left[ \begin{array}{cc}
                    1.919 & 0.112 \\
                    2.581 & -0.659 \\
                    1.912 & 1.679 \\
                    1.560 & -0.168
                    \end{array} \right].$ \\[1ex]

        \item DQN implementation details:
        \begin{enumerate}
            \item Stepsize: $\alpha_n = 2/n.$

            \item Exploration parameter $\epsilon = 0.05.$

            \item Replay buffer size: 10; Batch size: 8

            \item Target network update period: 8
        \end{enumerate}
    \end{itemize}

    \item Figure~\ref{fig:2PartitionOneSelfConsistent}

    \begin{itemize}
        \item MDP Details
        \begin{enumerate}
            \item $|\cS| = |\cA| = 2,$ i.e., 2 states and 2 actions.

            \item Transition matrix \\
            $\{\bP(s'|s, a_1)\}_{s, s'} = \left[ \begin{array}{cc}
                0.355 & 0.645 \\
                0.598 & 0.402
            \end{array} \right]$, \\[1ex]
    
            $\{\bP(s'|s, a_2)\}_{s, s'} = \left[ \begin{array}{cc}
                0.820 & 0.180 \\
                0.288 & 0.712
            \end{array} \right]$.

            \item Reward vector \\
            $r = \left[ \begin{array}{c}
                -0.599 \\
                -1.427 \\
                0.658 \\
                0.300
                \end{array} \right]$. \\

            \item Discount factor $\gamma = 0.75$.

        \end{enumerate}

        \item Feature matrix \\

         $\Phi = \left[ \begin{array}{cc}
    0.985 & 0.951 \\
    0.395 & 1.078 \\
    -0.904 & 1.276 \\
    0.063 & 1.214
    \end{array} \right]$. \\

    \item Exploration parameter $\epsilon = 0.1$.

    \end{itemize}
    
\end{enumerate}

%%%%%%%%%%%%%%%%%%%%%%%%%%%%%%%%%%%%%%%%%%%%%%%%%%%%%%%%%%%%

% % % %%%%%%%%%%%%%%%%%%%%%%%%%%%%%%%%%%%%%%%%%%%%%%%%%%%%%%%%%%%%

\section*{Acknowledgment}
We sincerely thank Vivek Borkar, Shalabh Bhatnagar, Konstantin Avrachenkov, Sean Meyn, Siva Theja Maguluri, Prashanth L. A., Alexandre Reiffers-Masson, and Kenny Young for their insightful discussions, which significantly contributed to the improvement of this paper.
%

% \section*{References}

\bibliographystyle{ieeetr}
\bibliography{references.bib}

\end{document}